\documentclass[11pt,twocolumn,letterpaper]{article}

\usepackage{wacv}

\usepackage{times}
\usepackage{latexsym}
\usepackage[utf8]{inputenc} 
\usepackage[T1]{fontenc}    
\usepackage{url}            
\usepackage{booktabs}       
\usepackage{amsfonts}       
\usepackage{nicefrac}       
\usepackage{microtype}      
\usepackage{graphicx}
\usepackage{amsmath}
\usepackage{hyperref}       
\usepackage{cleveref}       

\usepackage{xspace}
\usepackage{bm}
\usepackage{tabu}
\usepackage{xcolor}
\usepackage{caption}
\usepackage{subcaption}
\usepackage{multirow}
\usepackage{appendix}
\usepackage{etoolbox}
\usepackage{float}

\BeforeBeginEnvironment{appendices}{\clearpage}

 \hypersetup{
     colorlinks=false,
     linkcolor=blue,
     filecolor=blue,
     citecolor = black,      
     urlcolor=cyan,
     }

\title{Interpreting Transformed-based Models by Concept Classification}
\title{Finding Experts in Transformer Models}

\author{%
    Xavier Suau  \quad\qquad Luca Zappella  \quad\qquad Nicholas Apostoloff 
 \vspace{1mm}
\\
{\fontsize{11}{13}\texttt{\{xsuaucuadros, lzappella, napostoloff\}@apple.com}\selectfont}
  \vspace{1mm}
\\ 
  Apple 
\\
}

\makeatletter
\newcommand\footnoteref[1]{\protected@xdef\@thefnmark{\ref{#1}}\@footnotemark}
\makeatother

\newcommand{\fig}{Fig.~}
\newcommand{\tabref}{Table~}
\newcommand{\appref}{Appendix~}
\newcommand{\secref}{Sec.~}
\newcommand{\eq}{Eq.~}

\newcommand{\TM}{TMs\xspace}
\newcommand{\oneTM}{TM\xspace}
\newcommand{\ap}{\text{AP}}
\newcommand{\map}{\ap^{\star}}

\newcommand{\bx}{\bm{x}}
\newcommand{\bz}{\bm{z}}
\newcommand{\bu}{\bm{u}}
\newcommand{\by}{\bm{b}}

\newcommand{\np}{N^+_c}
\newcommand{\nm}{N^-_c}
\newcommand{\rr}{r^2}

\newcommand{\repr}{\bm{s}}
\newcommand{\overlap}[2]{\Omega(#1,#2)}

\newcommand{\gstar}{\gamma^{\star}}

\newcommand{\richallnopt}{\mathcal{X}_{\gamma}}
\newcommand{\richall}{\mathcal{X}_{\gamma}}

\newcommand{\richo}{\mathcal{X}_{\gamma}^{\text{homograph}}}
\newcommand{\richa}[2]{\mathcal{X}_{\gamma=#1}^{\text{#2}}}
\newcommand{\richstar}{\mathcal{X}_{\gstar}}

\newcommand{\wordnet}[1]{http://wordnetweb.princeton.edu/perl/webwn?s=#1\&sub=Search+WordNet\&o2=\&o0=1\&o8=1\&o1=1\&o7=\&o5=\&o9=\&o6=1\&o3=\&o4=\&h=0000000}

\newcommand{\gen}[2]{{\fontsize{#1}{11}\textsf{#2}\selectfont}}
\newcommand{\citep}[1]{\cite{#1}}

\wacvfinalcopy %

\def\wacvPaperID{979} %

\ifwacvfinal\fi
\begin{document}

\maketitle
\ifwacvfinal\thispagestyle{empty}\fi

\begin{abstract}
In this work we study the presence of expert units in pre-trained Transformer Models (\TM), and how they impact a model's performance. We define expert units to be neurons that are able to classify a concept with a given average precision, where a concept is represented by a binary set of sentences containing  the concept (or not). Leveraging the OneSec dataset \cite{Scarlini:ACL:2019}, we compile a dataset of 1641 concepts that allows diverse expert units in \TM to be discovered. We show that expert units are important in several ways: 
(1) The presence of expert units is correlated ($\rr=0.833$) with the generalization power of \TM, which allows ranking \TM without requiring fine-tuning on suites of downstream tasks. We further propose an empirical method to decide how accurate such experts should be to evaluate generalization. 
(2) The overlap of top experts between concepts provides a sensible way to quantify concept co-learning, which can be used for explainability of unknown concepts.
(3) We show how to self-condition off-the-shelf pre-trained language models to generate text with a given concept by forcing the top experts to be active, without requiring re-training the model or using additional parameters. 

\end{abstract}

\begin{figure}[t]
\begin{center}

{\fontsize{8}{11} \selectfont
\setlength{\tabcolsep}{1mm}
\begin{tabular}{p{0.93\columnwidth}l}
\toprule

\textsf{
\hspace{-0mm}{\color{gray}In this work we study the presence of expert units in pre-trained Transformer Models (TMs), and how they impact a model's performance. We define expert units to be neurons that are able to classify a specific concept with a given average precision,} hence being able to correctly classify a future ball's passing or shooting angle. We find that many multi-category TMs are specifically trained to indicate the accuracy of a given concept when following the instructions of a pre-specified skill coach and/or observers. In this experiment we test this hypothesis, using TMs that specialize in the shooting skills of throwing and catching.}\\


\bottomrule
\end{tabular}
}
  \caption{Example of generated text by GPT2-L conditioned on the WordNet concept \href{\wordnet{football}}{football\%1:04:00}{\protect\footnoteref{footwordnet}} ~by forcing only its top 50 expert units (0.012\% of the 414720 units analyzed) as determined by our interpretation method detailed in Section~\ref{sec:respo_units}. The beginning of this paper's abstract is given as context {\color{gray}(gray)}. 
Neither the interpretation nor the conditioning require re-training, fine-tuning or using additional parameters. 
Note the strong presence of concept \href{\wordnet{football}}{football\%1:04:00} in the generated text, including words like `coach' or `shooting'.  Even more interesting is how the the term `TM' appears in a sporting sense, and how TMs `specialize', taking the initial context also into account.\vspace{-2mm}
} 
  \label{tab:intro_sentence}
\end{center}
\end{figure}

\section{Introduction}
\label{sec:introduction}

Natural language processing (NLP) has evolved at a fast pace during recent years. Large and powerful models based on the Transformer architecture \citep{Vaswani:NIPS:2017} can now be trained on large datasets, achieving impressive performance on many NLP tasks, such as GLUE \citep{Wang:ICLR:2019} or SQuAD \citep{Rajpurkar:EMNLP:2016,Rajpurkar:ACL:2018}. Training \TM is tedious due to both the size of models and datasets, and requires resources that are unavailable to many users. For example, in \citep{Nvidia:GITHUB:2019} the model was trained on 512 GPUs. A recent trend is to pre-train  these large models on diverse datasets and make them available to the community \citep{Howard:ACL:2018,Devlin:NAACL:2018,Radford:ARXIV:2019,Lample:ARXIV:2019,Yang:NEURIPS:2019,liu:ARXIV:2019,Sanh:ARXIV:2019}, so that end-users can leverage the powerful features learnt to solve downstream tasks. 


However, it is not fully understood why these models perform so well. Inspired by observations in neuroscience stating that the human brain is a network of hierarchically organized specialized modules \citep{Hill:2006} and that neurons become more selective as humans learn \citep{Najafi:Neuron:2019}, we aim to find specialized modules in \TM, the expert units. \textit{We hypothesize that the presence of expert units is related to the knowledge acquired by \TM and to their performance.}

Previous work in image processing has shown that CNNs and GANs learn representations of specific objects at a filter level \citep{Bau:CVPR:2017,Bau:ICLR:2019} and by filter combinations  \citep{Fong:CVPR:2018}, even if those objects were never explicitly labeled during training. The key idea behind these works is to consider CNN feature maps as segmentation masks, which allows quantifying the coherence with a densely labeled image by means of intersection over union (IOU). These works have also inspired our research, however there are fundamental differences in our work: (1) In NLP it is harder to  define a concept with a single sentence, thus we propose to represent concepts with sets of positive and negative sentences as explained in \secref\ref{sec:sentenceconcepts}. We collect a total of 1641 concepts, leveraging the OneSec dataset \citep{Scarlini:ACL:2019}. (2) We consider the most basic units in \TM, the neurons, as expert unit candidates, which allows computing average precision $\ap$ (\ie area under the precision-recall curve) to quantify how a unit is able to disambiguate a concept	. (3) Since sentences can be of arbitrary length, we maxpool the unit responses in time to be invariant to length. The proposed method to identify and rank experts is detailed in \secref\ref{sec:respo_units}.

In \secref\ref{sec:concept_richness} we define a metric called \textit{concept expertise}, and we show that it is strongly correlated ($\rr=0.833$) with the generalization power of \TM. As a measure for model generalization we use the average performance on diverse downstream tasks: all GLUE tasks + SQuAD v1.1/2.0. 
We propose an empirical method to compute the optimal expertise level that maximizes the correlation between generalization and expertise. 
The obtained expertise ($\ap$ above 0.985) shows that generalization is related to the presence of extremely good and diverse experts. Our definition of expertise enables the ranking of \TM withoutfine-tuning on large suites of downstream tasks (current practice), mitigating the need for hyper-parameter search and the problem of downstream task bias \citep{Niven:ACL:2019}. 
Moreover, the proposed concept dataset can be easily enriched for finer model comparison.

In \secref\ref{sec:concept_repr} we show that concepts with similar meaning are co-learnt by a certain number of experts. We define concept overlap to quantify co-learning, and we show its utility for concept explainability. 

The presence of experts is also exploited in \secref\ref{sec:forcing} to self-condition a pre-trained language model (LM) to generate text that contains a specific concept (see \fig\ref{tab:intro_sentence} for an example). We base our approach on the product of experts formulation introduced by \citep{Hinton:ICANN:1999} and adapted to image generative models by \citep{Nguyen:CVPR:2017} by training an external conditional expert. 
In addition to applying the product of experts formulation to a new domain (NLP) and new architecture (\oneTM), a notable difference
is that \textit{we consider that conditional experts already exist in the pre-trained model}. The results in \secref\ref{sec:generation} show that only a small number of experts is required to induce a concept in the model output, supporting our hypothesis. Other works have tackled LM conditioning \citep{Hu:ICML:2017,Romanov:NAACL:2019,Chen:NAACL:2019,Keskar:Arxiv:2019}, all based on learning disentangled concepts during training. To the best of our knowledge, our work is the first to condition an off-the-shelf pre-trained LM without fine-tuning, re-training or using additional parameters.  Furthermore, our method is extremely simple to implement. 

\section{Related work}
\label{sec:soa}

The NLP community is experiencing a sharp increase in interpretation methods. We focus on those exploring Transformer architectures, which are the keystone for most of the recent top performing models.


\paragraph{Saliency} Some works focus on analyzing the self-attention layers in the Transformer blocks, visualizing saliency \citep{Ghaeini:EMNLP:2018} or studying how attention heads attend to different word families \citep{Clark:ACL:2018}. The analysis of attention layers usually results in a word-word relationship, which can make it hard to extract model-wide conclusions. Moreover, recent studies show that saliency based methods may be invariant to the model or the data \citep{Adebayo:NeurIPS:2018} and can be easily manipulated \citep{Dombrowski:NeurIPS:2019}.


\paragraph{Intermediate discriminators} Another trend is to probe the model with a dataset representative of some downstream task, either at a sentence level \citep{Adi:ICLR:2017,Conneau:ACL:2018} or at a word level \citep{Tenney:ICLR:2019,Liu:NAACL:2019}. The common practice is to train a classifier on top of selected intermediate features to assess their discriminative power. These approaches inspired our work, but rather than learning classifiers to solve downstream tasks, we probe the \oneTM's responses directly with a large set of concepts unrelated to the final task. We propose treating the units of an already trained \oneTM as classifiers themselves. 


\paragraph{Disentangled learning} Most methods tackling concept learning are based on \textit{training dedicated architectures}. Concepts such as syntax and semantics \citep{Chen:NAACL:2019}, meaning and form \citep{Romanov:NAACL:2019}, or sentiment and tense \citep{Hu:ICML:2017} can be disentangled by capturing different intrinsic aspects of text. Although effective, these methods suffer from the requirements of \oneTM training. Our approach does not require training or knowledge of the training procedure. It requires only a pre-trained model and a dataset of concepts, such as the dataset described in \secref\ref{sec:sentenceconcepts}.

\section{SentenceConcepts: A dataset of concepts represented by sentences}
\label{sec:sentenceconcepts}
We propose a data-driven approach to describe a concept. We collect $\np$ positive sentences that contain concept $c$ and $\nm$ negative sentences that do \emph{not} contain concept $c$. Such flexible definition allows diverse types of concepts to be represented. For example, one can collect positive sentences containing a keyword with a specific meaning, \eg \emph{note} as a reminder, or \emph{note} as a musical frequency. 
We construct our dataset leveraging the recently published OneSec dataset \citep{Scarlini:ACL:2019}, which contains sentences with one keyword annotated with a WordNet\footnote{\label{footwordnet}We adopt the WordNet sense key formulation, of the form \texttt{lemma\%A:BB:CC}, clickable as web links across the paper.} sense  \citep{Princeton:WordNet:2020}. We consider 2 concept categories: 
\begin{itemize}
\item \textbf{Sense}: Positive sentences contain a keyword with a specific WordNet sense, whereas negative sentences do not contain the keyword.
\item \textbf{Homograph}: Positive sentences contain a keyword with a specific WordNet sense, whereas negative sentences contain the same keyword with a different meaning. Intuitively, \textit{homograph} concepts are harder to disambiguate than \textit{sense} concepts.
\end{itemize}
In total, the dataset contains 1344 \textit{sense} and 297 \textit{homograph} concepts. The complete list of concepts and details on the annotations are provided in \appref\ref{app:concept_list}.
The number of sentences collected is constrained by availability in the source dataset. We limit the data per concept to $\np, \nm \in [100, 1000]$, randomly sampling when more than 1000 sentences are available.

\section{Expert Units}
\subsection{Finding expert units}
\label{sec:respo_units}

\begin{figure}[t]
  \centering
\begin{subfigure}[b]{0.4\textwidth}
    \includegraphics[width=\textwidth]{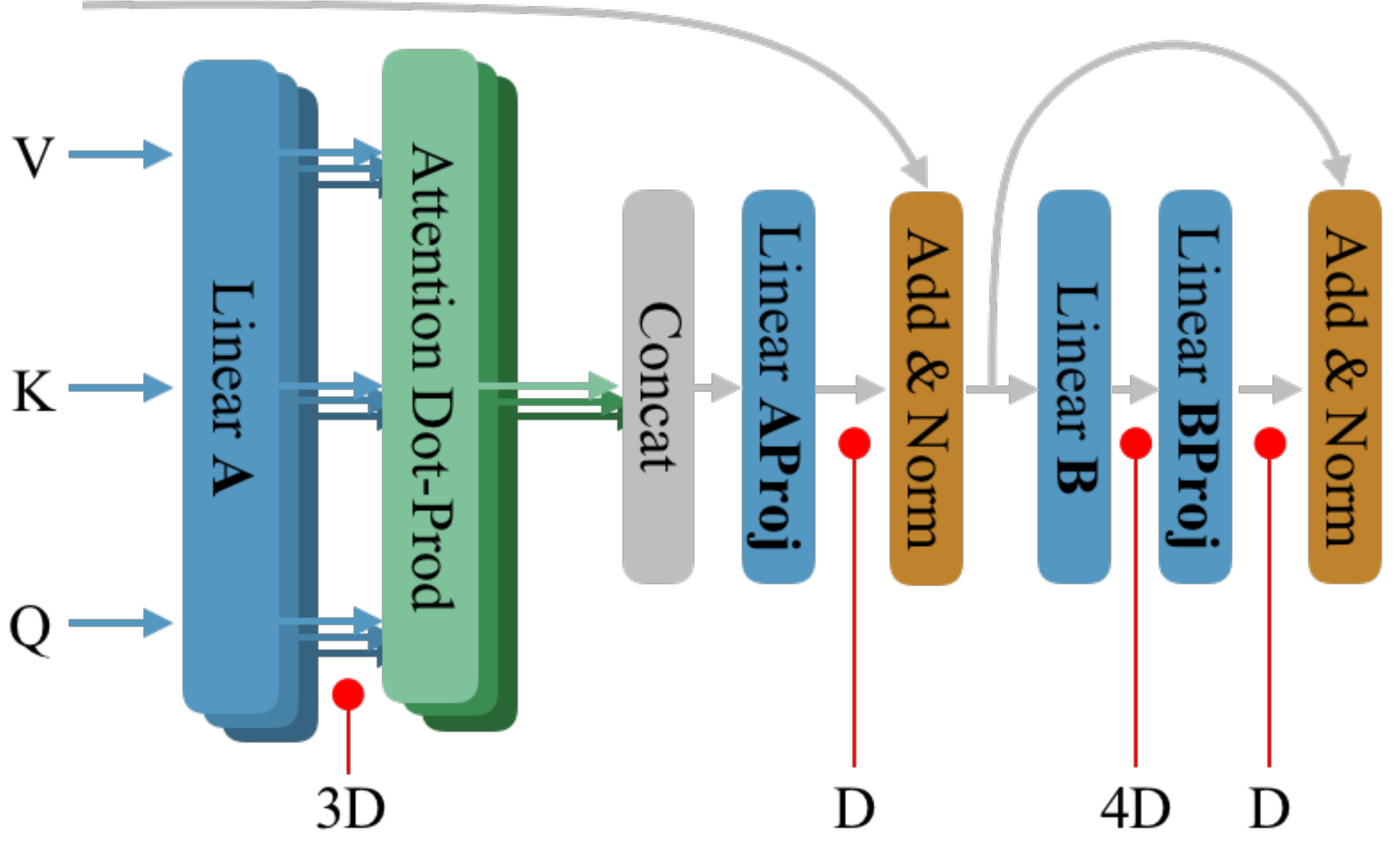}
  \end{subfigure}
  \caption{Schema of a Transformer block \citep{Vaswani:NIPS:2017}. In this work we analyze the units in the linear layers A, Aproj, B and Bproj in each block (red dots), where $D$ is the dimensionality of the embedding. For example, in GPT2-large ($D=1280$ and 36 blocks) we analyze $36\cdot 9D=414720$ units. \vspace{-2mm}}\label{fig:transformer_schema} 
\end{figure}

Let $\bx_i = [\bx_{i, 1}, \cdots, \bx_{i, T}]\in \mathbb{R}^{D\times T}$ be a sentence composed of an arbitrary number $T$ of tokens $\bx_{i,t}\in\mathbb{R}^D$, with $D$ being the dimensionality of the token embedding. A layer $\ell$ of a \oneTM produces an intermediate representation $\bz_{i}^{\ell} = [\bz_{i, 1}^{\ell}, \cdots, \bz_{i, T}^{\ell}] \in \mathbb{R}^{D^{\ell}\times T}$, where typically $D^{\ell}$ is a multiple of $D$. Let $[u_i^{(\ell, 1)}, \cdots, u_i^{(\ell, D^{\ell})}]^\top=\text{maxpool}\big(\bz_{i}^{\ell}, \text{axis}=1\big) \in\mathbb{R}^{D^{\ell}}$ be the intermediate representation max-pooled in the temporal dimension, where each element $u_{i}^{(\ell, k)}\in\mathbb{R}$ is the response of unit $k$ in layer $\ell$ to sentence $i$. 
For simplicity the $(\ell, k)$ indexing is replaced by $m=1..M$, with $M$ being the total number of units in the model. The layers analyzed are shown in \fig\ref{fig:transformer_schema}. Let $\bu_c^m \in \mathbb{R}^{N_c}$ (where $N_c=\np+\nm$) be the pooled response of unit $m$ 
to the sentences that represent concept $c$, and let $\by_c \in \mathbb{Z}_2^{N_c}$ be the binary labels for such sentences.  We treat a unit as a binary classifier for the input sentences, and consider the whole network as a collection of binary classifiers. By using $\bu_c^m \in \mathbb{R}^{N_c}$ as prediction scores for $\by_c$, we can compute $\ap_c^m = \ap(\bu_c^m, \by_c) \in [0, 1]$ per unit $m$ and per concept $c$, which allows ranking units by expertise (or $\ap$) on each concept. We expect that, given the large search space, certain classifiers will perform well on specific concepts: the expert units.

\begin{figure*}[ht]
  \centering
\begin{subfigure}[b]{\textwidth}
    \includegraphics[width=0.99\textwidth]{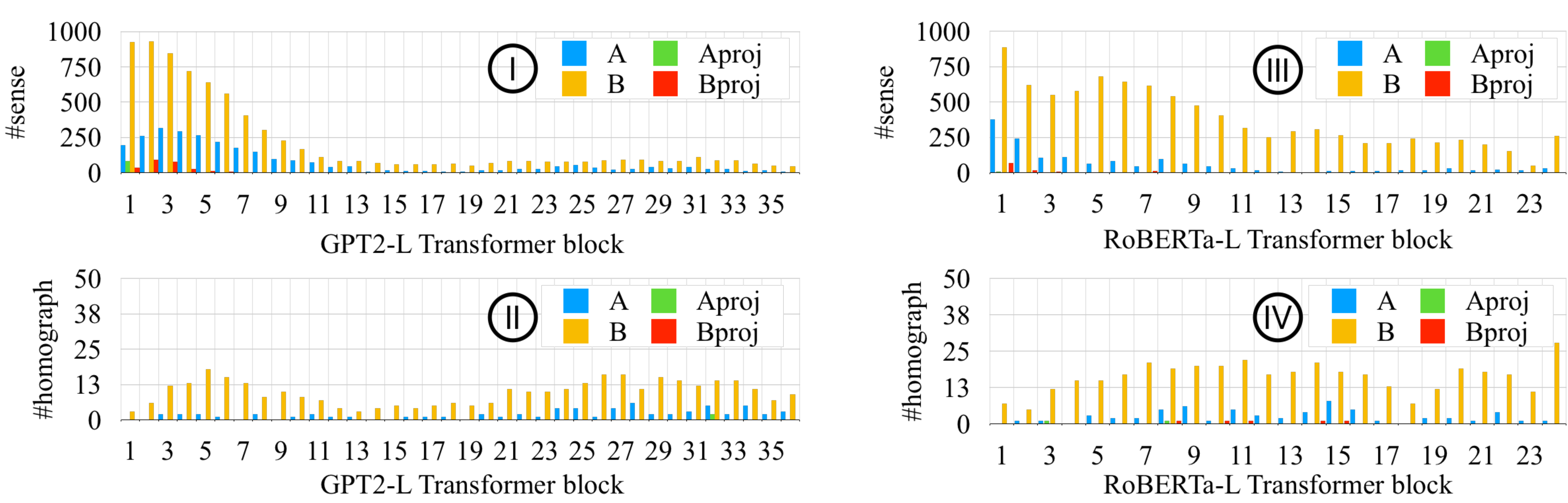}
  \end{subfigure}
    \caption{
  Number of acquired concepts $(\gamma = 0.95)$ per layer type  for model  GPT2-L (I, II) and  RoBERTa-L (III, IV). Shallow layers acquire more \textit{sense} concepts than deep layers, such effect being emphasized in GPT2-L. This behavior is similar to the one observed by \citep{Bau:ICLR:2019} for image GANs. We further observe that B layers acquire 3.5x more concepts than A layers. Aproj and Bproj layers  acquire very few concepts, suggesting that the expanding layers (A, B) in the Transformer block (\fig\ref{fig:transformer_schema}) are more prone to learn concepts. \textit{Homograph} concepts (II, IV) are spread across the layers,  few being detected in the first Transformer blocks. \vspace{-2mm} }\label{fig:concept_distribution}
\end{figure*}

\subsection{Concept expertise $\mathcal{X}_{\gamma}$}
\label{sec:concept_richness}
Let $\map_c = \underset{m}\max\big\{ \ap_c^m \}$ be the $\ap$ of the best expert for concept $c$. Let $\gamma$ be the acquisition threshold so that a concept is considered as \emph{acquired} in the model if $\exists\, \ap_c^m \geq \gamma\; \forall m$ (or $\map_c \geq \gamma$). We  define \textit{concept expertise} as the percentage of concepts acquired by the model:
\vspace{-2mm}
\begin{equation}
\label{eq:richness}
\mathcal{X}_{\gamma} = \frac{ |\{c \;\; \text{s.t.} \;\; \map_c \geq \gamma\}| }{ |C| }\quad \forall\,c\in C.
\end{equation}

\paragraph{Finding an optimal $\gstar$ value}
\label{sec:gamma}
The choice of $\gamma$ is important to compute the concept expertise $\richallnopt$ in \eq\eqref{eq:richness}. The goal is to obtain an optimal $\gstar$ that produces an expertise  representative of the generalization power of \TM. As a measure of generalization, we use the average performance of each model on typical downstream tasks: 
the 10 datasets composing GLUE with their different reported metrics \citep{Wang:ICLR:2019}
and SQuAD v1.1/2.0 \citep{Rajpurkar:EMNLP:2016,Rajpurkar:ACL:2018}. The reported performance is presented in Table~\ref{tab:downstream} in \appref\ref{app:downstream}. 

We measure the squared Pearson's correlation coefficient $\rr$ between $\richall$ and generalization. The obtained $\gstar$ tells the level of expertise required for a concept to be considered as acquired.
We then define the optimal value of $\gamma$ as 
\vspace{-2mm}
\begin{equation}
\label{eq:optgamma}
\gstar = \underset{\gamma}{\text{argmax}}\Big(  \frac{1}{N_{\text{tasks}}}  \sum_{\text{task}}   \rr(\richallnopt, \text{task})      \Big),
\end{equation}
with $\gamma \in [0.5, 1)$. To assess the robustness of $\gstar$, the tasks are randomly split into reference and test sets, with a ratio 60/40\%. Next, we compute $\gstar$ for each subset, and we measure the RMSE between the obtained values on the reference and test set (10 random splits). We treat the \textit{sense} and \textit{homograph} concepts independently since they are fundamentally different. We obtain a $\gstar = 0.997$ with a RMSE of 0.0004 for concepts \textit{sense} and $\gstar = 0.985$ with a RMSE of 0.0028 for concepts \textit{homograph}. The low RMSE shows that the value of $\gstar$ generalizes well on disjoint sets of tasks. For simplicity, we express the optimal values as $\gstar = \{\text{sense: } 0.997, \text{homograph: } 0.985\}$, and we define the combined expertise as 

\vspace{-4mm}
\begin{equation}
\label{eq:comb_rich}
\richstar = \frac{1}{N_{\text{concepts}}} \sum_{g \in (\text{sense, homograph})} \hspace{-8mm}N_g \richa{\gstar[g]}{g}.
\end{equation}
The high values of $\gstar$, together with the obtained $\rr=0.833$, suggest that the number of good and diverse experts in the model is correlated with its generalization power (see \tabref\ref{tab:correlation_datasets} for full results).

\subsection{Results on expert units}
\label{sec:results}

All of the pre-trained models evaluated are obtained from the \href{https://huggingface.co/transformers/}{Huggingface Transformers repository} \citep{Wolf:ARXIV:2019}, version 2.1.1.
More precisely, we analyze\footnote{The corresponding names in the transformers repository are: \texttt{bert-base/large-cased, roberta-base/large/large-mnli, distilbert-base-uncased, gpt2-$\emptyset$/medium/large} and \texttt{xlm-mlm-en-2048}.}: BERT-B/L \citep{Devlin:NAACL:2018}, RoBERTa-S/L/Lm \citep{Liu:NAACL:2019}, DistilBERT \citep{Sanh:ARXIV:2019}, GPT2-S/M/L \citep{Radford:ARXIV:2019} and XLM \citep{Lample:ARXIV:2019}. The sentences are tokenized using the default settings in the repository.

\subsubsection{Concept distribution results}

The distribution of acquired concepts per layer type is shown in \fig\ref{fig:concept_distribution}, for models GPT2-L (I, II) and RoBERTa-L (III, IV).
A concept is considered acquired in a layer  $\ell$ if $\exists\, \ap_c^{(k,\ell)} \geq \gamma$. 
In this experiment, we use $\gamma=0.95$ for visualization purposes, $\gstar$ being too restrictive.

 We observe that shallow layers in \TM accumulate more concepts than deep layers. Within the Transformer blocks (see \fig\ref{fig:transformer_schema}) in GPT2-L, B layers acquire about 3.5x more concepts than A layers and more than 10x than Aproj and Bproj layers. This suggests that the expanding layers (A and B) in the Transformer block are better at learning concepts at a unit level. RoBERTa-L produces a similar distribution of concepts, with A and B layers accumulating most of the concepts. Compared to GPT2-L, RoBERTa-L has a smaller drop in the number of concepts from shallow to deep layers. GPT2-L is a generative model composed of Transformer decoders, while RoBERTa is a stack of encoders. Our results show that generative architectures in NLP tend to accumulate concepts early in the model. Such an observation was reported  by \citep{Bau:ICLR:2019} in the image GAN domain, but to the best of our knowledge we report the first observation of this phenomenon in the NLP domain. 
Refer to \appref\ref{app:concept_distribution} for results on other models.


\subsubsection{Expertise and generalization results}
\label{sec:downstream}


\begin{table}[t]
  \centering
{\fontsize{8}{10} \selectfont
\setlength{\tabcolsep}{1mm}
\begin{tabular}{lccc|c}
\toprule
Model & Model size &   $\richa{0.997}{sense}$ & $\richa{0.985}{homograph}$ &   $\richstar$ \\
\midrule
BERT-B          &       110M &   1.04\% (14) &   5.72\% (17) &   1.89\% \\
BERT-L           &       330M &   7.51\% (101) &   5.72\% (17)&   7.19\% \\
Distilbert    &        66M &   3.65\% (49)&   5.72\% (17)&   4.02\% \\
GPT2-S                       &       117M &   1.79\% (24) &   1.35\% (4)&   1.71\% \\
GPT2-M                &       345M &   3.65\% (49) &   3.03\% (9)&   3.53\% \\
GPT2-L                &       774M &  15.03\% (202) &   3.37\% (10)&  12.92\% \\
RoBERTa-B               &       125M &   1.71\% (23)&   3.70\% (11)&   2.07\% \\
RoBERTa-L              &       355M &  14.66\% (197) &   5.05\% (15)&  12.92\% \\
RoBERTa-Lm         &       355M &  17.86\% (240) &   4.04\% (12)& \textbf{ 15.36\%} \\
XLM            &       667M &   9.30\% (125)&   5.39\% (16) &   8.59\% \\
\bottomrule
\end{tabular}
}

  \caption{Expertise for \textit{sense} and \textit{homograph} concepts, and combined expertise $\richstar$ \eqref{eq:comb_rich}.
  In parenthesis the actual number of concepts acquired. RoBERTa-Lm shows the highest $\richstar = 15.36\%$. The models analyzed obtain a low \textit{homograph} expertise $\richo \leq 5.72\%$ compared to \textit{sense} concepts. \vspace{-2mm}}
  \label{tab:learnt}
\end{table}

\paragraph{Concept expertise}
The concept expertise obtained by the considered models is summarized in Table~\ref{tab:learnt}. RoBERTa-Lm is the model that achieves better combined expertise $\richstar = 15.36\%$, followed by RoBERTa-L and GPT2-L, both with 12.92\%. 
It is interesting to note that RoBERTa-L doubles the concept expertise of BERT only by modifying the training procedure and the data. 

\begin{figure}[tb]
  \centering
  \begin{subfigure}[b]{\columnwidth}
    \includegraphics[width=\textwidth]{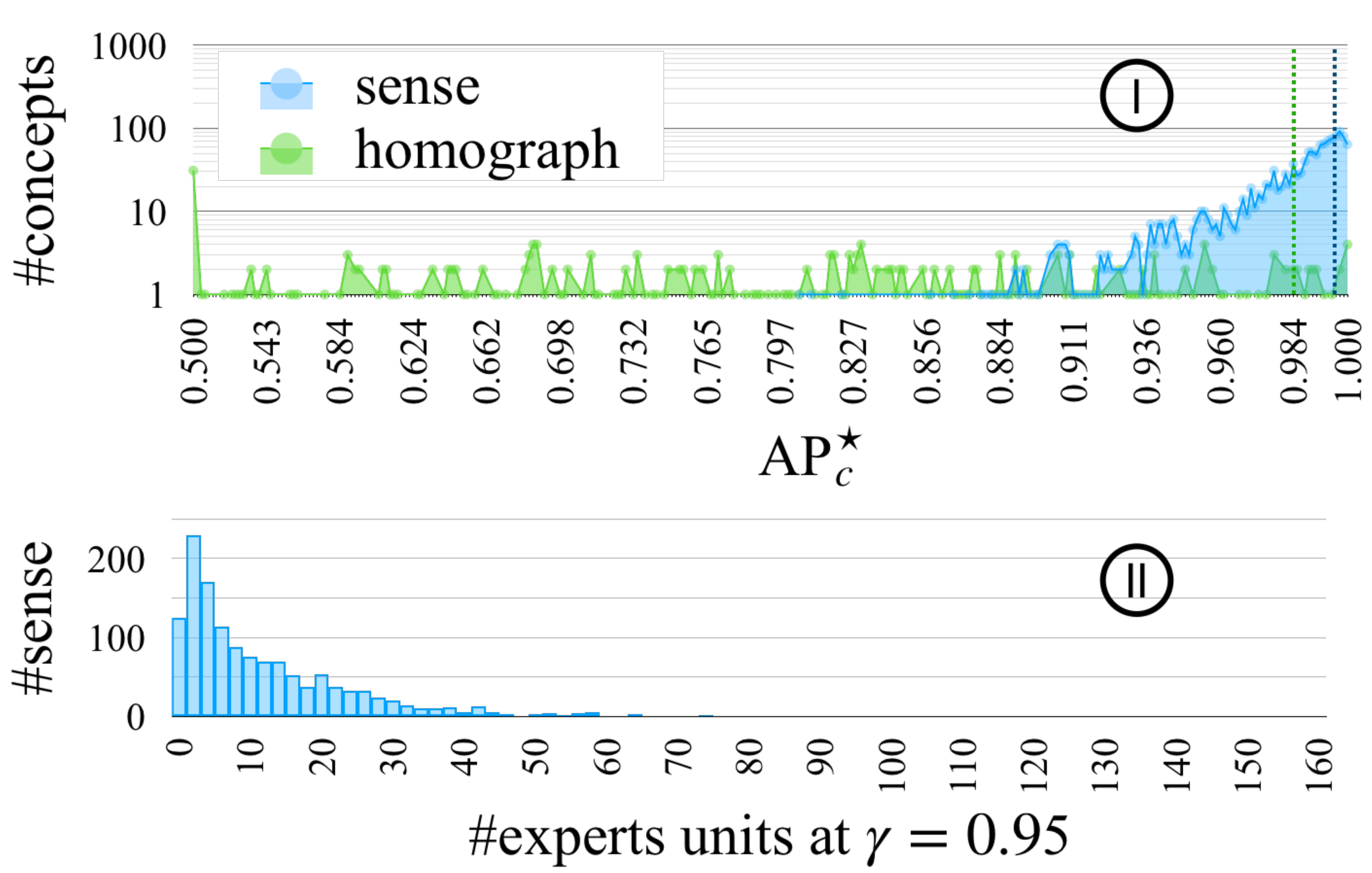}
  \end{subfigure}
  \caption{
(I) Histogram of the $\map_c$ per concept for model RoBERTa-L. Most of \textit{sense} concepts are detected with $\map_c > 0.90$, while \textit{homograph} concepts present a wide range of $\map_c$. The vertical lines correspond to the values $\gstar$ found in \secref\ref{sec:gamma}. (II) Histogram of the number of expert units that acquire a \emph{sense} concept at $\gamma = 0.95$. Most of the concepts have less than 50 experts associated, a very low value (0.022\%) compared to the 221184 units analyzed for RoBERTa-L showing that the number of experts per concept in a \oneTM is very selective. \vspace{-2mm}}\label{fig:histograms}
\end{figure}

We observe that \textit{sense} concepts are acquired better than \textit{homograph} concepts, as expected given the difficult disambiguation of the latter.  In \fig\ref{fig:histograms}.(I) we show the histogram of $\map_c$ for all concepts. Note how many \textit{homograph} concepts are not being detected, while almost all \textit{sense} concepts are detected with $\map_c>0.90$.  Building pre-trained models inherently able to disambiguate \textit{homograph} concepts at unit level remains a challenge, and we speculate that such knowledge will help the models generalize even better.

In \fig\ref{fig:histograms}.(II) we show the histogram of expert units that acquire  a \textit{sense} concept at $\gamma = 0.95$, for model RoBERTa-L. We observe that most of the concepts have less than 50 dedicated expert units (0.022\%), with a median of 7 experts (0.0032\%) per concept. Taking into account that 221184 units were analyzed for this model, we conclude that \TM dedicate very specific groups of experts to different concepts. The results in \secref\ref{sec:generation} how that these experts are causal.

%

\paragraph{Generalization}

In Table~\ref{tab:correlation_datasets}, we show that concept expertise $\richstar$ in \eq\eqref{eq:comb_rich} is a robust measure of generalization of \TM, and can be used as a model evaluation metric rather than fine-tuning on downstream tasks. We report the squared Pearson's correlation coefficient $\rr \in [0, 1]$ of the linear regression between the performance of \TM on downstream tasks (see \secref\ref{sec:gamma}) versus $\richstar$ . Only the models reporting results are used in the correlation analysis. For comparison, we report in column (T) the average correlation between performance on a task with all the other tasks, as well as the correlation of downstream performance with the model size in column (S). 
We observe that $\richstar$ is strongly correlated with the performance on downstream tasks ($\rr>0.833$), which is higher than the average correlation among tasks (0.826), the latter requiring evaluation (involving fine-tuning) on all the tasks. The correlation with $\richstar$ is better than the correlation between tasks for 12 of the 16 metrics used. 

We further see that model size is not a metric that generalizes well, since it correlates strongly with the perfromance on some tasks (\eg SQuAD), but weakly with other tasks (\eg QNLI or MRPC).
For all of the tasks\footnote{With the exception of STS-(p), which shows 0.0 correlation with model size and a very poor correlation (0.49) with the other tasks too.} where the correlation with model size $\rr<0.3$, we obtain a correlation with $\richstar$  of $\rr>0.75$, reinforcing the claim that concept expertise  $\richstar$ is a good measure of generalization.

\begin{table}[tb]
  \centering
{\fontsize{8}{8.5} \selectfont
\setlength{\tabcolsep}{2mm}
\begin{tabular}{l|lrrr}
\toprule
Models used & Task &  (S)  &  (T)  &  $\richstar$  \\
\midrule
\multirow{12}{*}[-1.5ex]{\shortstack[l]{BERT-B/L\\Distilbert\\RoBERTa-L\\XLM}}
& GLUE Score     &       0.361 &                0.871 &         \textbf{0.892} \\
& CoLA           &       0.572 &                0.821 &         \textbf{0.874} \\
& SST-2          &       0.554 &                0.849 &         \textbf{0.864} \\
& MRPC (acc)     &       0.163 &                0.714 &         \textbf{0.753} \\
& MRPC (F1)      &       0.162 &                \textbf{0.916} &         0.855 \\
& STS-B (p)      &       0.000 &                \textbf{0.490} &         0.401 \\
& STS-B (s)      &       0.115 &                \textbf{0.903} &         0.840 \\
& QQP (acc)      &       0.340 &                0.936 &         \textbf{0.944} \\
& QQP (F1)       &       0.397 &                0.834 &         \textbf{0.859} \\
& MNLI-m         &       0.603 &                \textbf{0.833} &         0.771 \\
& MNLI-mm        &       0.452 &                0.906 &         \textbf{0.944} \\
& QNLI           &       0.286 &                0.857 &         \textbf{0.923} \\
& RTE            &       0.332 &                0.861 &         \textbf{0.873} \\
& WNLI           &       0.314 &                \textbf{0.751} &         0.619 \\
& AX             &       0.426 &                0.914 &         \textbf{0.956} \\
\midrule
\multirow{4}{*}[0px]{\shortstack[l]{BERT-B/L\\DistilBERT\\RoBERTa-L}}
 &&&&\\
& SQuAD 1.1 (F1) &       \textbf{0.899} &                0.840 &         0.850 \\
& SQuAD 2.0 (F1) &       \textbf{0.961} &                0.752 &         0.937 \\
&&&&\\
\midrule
& Average            &       0.408 &                0.826 &         \textbf{0.833} \\
\bottomrule
\end{tabular}
}
  \caption{Squared Pearson's coefficient $(\rr)$ of the linear regressions between the reported performance on various tasks versus the model size (S),  the average versus all the other tasks (T) and the combined concept expertise $\richstar$ (\eq\eqref{eq:comb_rich}). Only those models reporting results on each dataset are used (first column). 
  On average, the correlation using $\richstar$ (0.833) is better than the average correlation among all tasks (0.826), the latter requires evaluation (involving fine-tuning) on all the tasks. 
  \vspace{-2mm}} 
  \label{tab:correlation_datasets}
\end{table}

\section{Concept overlap $\Omega$}
\label{sec:concept_repr}

Recent works in image processing have shown that CNN filters can represent multiple concepts \citep{Bau:CVPR:2017,Bau:ICLR:2019,Fong:CVPR:2018}. Based on this observation, we propose a method to explore co-learning of concepts by \TM in the NLP domain. We first set a threshold  $\tau_c = \text{percentile}_{99}(\ap_c^m)$ per concept, then we define $\repr_c = \{1 \;\; \text{if}\;\; \ap_c^m > \tau_c\;\; \text{else}\;\;0\} \in \mathbb{Z}_2^M$ as our binary concept representation. Note that $\repr_c$ has elements with value 1 for the top 1\% units classifying the concept. Let the overlap between concepts $q$ and $v$ be
\vspace{-2mm}
\begin{equation}
\label{eq:overlap}
\overlap{q}{v} = \frac{\|\repr_q \cap \repr_v\|_1 }{\|\repr_q \cup \repr_v\|_1 } \in [0, 1],
\end{equation}
representing the number of top units that classify both concepts with high $\ap_c^m$.


\begin{table}[t]
  \centering
{\fontsize{8}{10} \selectfont
\setlength{\tabcolsep}{1mm}
\begin{tabular}{llc}
\toprule
\shortstack[l]{Query definition} & {Concept} &  $\overlap{{\color{orange}q}}{v}$ \\
\midrule
\multirow{5}{*}[0px]{\shortstack[l]{A seat for one\\person, with a support\\for the back.}}
& {\color{orange}\href{http://wordnetweb.princeton.edu/perl/webwn?s=chair&sub=Search+WordNet&o2=&o0=1&o8=1&o1=1&o7=&o5=&o9=&o6=1&o3=&o4=&h=000000000000}{chair\%1:06:00} (query)}                 &    1.000 \\
& \href{http://wordnetweb.princeton.edu/perl/webwn?o2=&o0=1&o8=1&o1=1&o7=&o5=&o9=&o6=1&o3=&o4=&s=table&h=00000000&j=1#c}{table\%1:06:01}                  &    0.458 \\
& \href{http://wordnetweb.princeton.edu/perl/webwn?s=bed&sub=Search+WordNet&o2=&o0=1&o8=1&o1=1&o7=&o5=&o9=&o6=1&o3=&o4=&h=00000000}{bed\%1:06:00}                     &    0.361 \\
& \href{http://wordnetweb.princeton.edu/perl/webwn?s=cup&sub=Search+WordNet&o2=&o0=1&o8=1&o1=1&o7=&o5=&o9=&o6=1&o3=&o4=&h=0000000000000}{cup\%1:06:00}                     &    0.341 \\
& \href{http://wordnetweb.princeton.edu/perl/webwn?s=table&sub=Search+WordNet&o2=&o0=1&o8=1&o1=1&o7=&o5=&o9=&o6=1&o3=&o4=&h=00000000000}{table\%1:06:01} VS. \href{http://wordnetweb.princeton.edu/perl/webwn?o2=&o0=1&o8=1&o1=1&o7=&o5=&o9=&o6=1&o3=&o4=&s=table&h=00000000&j=1#c}{table\%1:14:00} &    0.336 \\
& \href{http://wordnetweb.princeton.edu/perl/webwn?s=floor&sub=Search+WordNet&o2=&o0=1&o8=1&o1=1&o7=&o5=&o9=&o6=1&o3=&o4=&h=00000000}{floor\%1:06:00}                   &    0.328 \\
\midrule
\multirow{5}{*}[0px]{\shortstack[l]{The position\\of professor.}}
& {\color{orange}\href{http://wordnetweb.princeton.edu/perl/webwn?s=chair&sub=Search+WordNet&o2=&o0=1&o8=1&o1=1&o7=&o5=&o9=&o6=1&o3=&o4=&h=000000000000}{chair\%1:04:00} (query)}                  &    1.000 \\
& \href{http://wordnetweb.princeton.edu/perl/webwn?s=chair&sub=Search+WordNet&o2=&o0=1&o8=1&o1=1&o7=&o5=&o9=&o6=1&o3=&o4=&h=000000000000}{chair\%1:04:00} VS. \href{http://wordnetweb.princeton.edu/perl/webwn?s=chair&sub=Search+WordNet&o2=&o0=1&o8=1&o1=1&o7=&o5=&o9=&o6=1&o3=&o4=&h=000000000000}{chair\%1:06:00} &    0.575 \\
& \href{http://wordnetweb.princeton.edu/perl/webwn?s=fellow&sub=Search+WordNet&o2=&o0=1&o8=1&o1=1&o7=&o5=&o9=&o6=1&o3=&o4=&h=0000000}{fellow\%1:18:02}                  &    0.371 \\
& \href{http://wordnetweb.princeton.edu/perl/webwn?s=director&sub=Search+WordNet&o2=&o0=1&o8=1&o1=1&o7=&o5=&o9=&o6=1&o3=&o4=&h=0000000}{director\%1:18:03}                &    0.297 \\
& \href{http://wordnetweb.princeton.edu/perl/webwn?s=administration&sub=Search+WordNet&o2=&o0=1&o8=1&o1=1&o7=&o5=&o9=&o6=1&o3=&o4=&h=00000}{administration\%1:04:00}          &    0.243 \\
& \href{http://wordnetweb.princeton.edu/perl/webwn?s=member&sub=Search+WordNet&o2=&o0=1&o8=1&o1=1&o7=&o5=&o9=&o6=1&o3=&o4=&h=000000}{member\%1:18:00}                  &    0.241 \\
\bottomrule
\end{tabular}
}
 \caption{
Top-5 concepts in terms of expert overlap $\overlap{{\color{orange}q}}{v}$ (\eq\eqref{eq:overlap}) with a {\color{orange}{query}} concept for RoBERTa-L. The overlap shows the amount of top experts shared by two concepts ${\color{orange}q}$ and $v$. Even if the word representing the concept is the same (chair), the concepts overlapping are different and relate to the actual WordNet definition (click each concept for WordNet link), showing that the model takes the meaning into account. Concepts marked with `VS.' are \textit{homograph} concepts.
 \vspace{-2mm}}
  \label{tab:knn_chair}
\end{table}

\subsection{Concept overlap results}
\label{sec:colearning_results}
In \tabref\ref{tab:knn_chair}, we show that concepts with related senses present a high overlap of top experts, evidencing concept co-learning. 
More precisely,  we show the 5 concepts $v$ with highest overlap $\overlap{q}{v}$ (defined in \eq\eqref{eq:overlap}) with a query concept $q$. The query concepts considered are represented by the same word (chair) but with different WordNet sense (first row of \tabref\ref{tab:knn_chair}). Observe how the top-5 concepts are coherent with the definition of the query. The fact that the \textit{homograph} concept that disambiguates the query appears with high overlap is also interesting. Concept co-learning can be used for explainability: given a concept with unknown definition (\eg failure cases in a dialogue system), the overlapping concepts can help explain it.
See \appref\ref{app:co_learning} for more results including t-SNE projections \citep{VanDerMaaten:JMLR:2008} of $\repr_c$.

\begin{table*}[t]
  \centering
{\fontsize{8.5}{8.5} \selectfont
\setlength{\tabcolsep}{1mm}
\begin{tabular}{lp{13.5cm}}
\toprule
K forced &        \hspace{-0mm}{\color{gray}Once upon a time}   {\small$+$} Generated induced for concept \href{\wordnet{bird}}{bird\%1:05:00} (warm-blooded egg-laying vertebrates) \\
\midrule
0 ($0\%$)   & \gen{6.8}{, I had a friend who used to teach high school English and he was like, "Oh, all you have to do is just get out there}  \\
\midrule
40 ($0.009\%$)  & \gen{6.8}{ , many of these treasures were worth hundreds of thousands of dollars.\textbackslash n But this isn't the first time that a horse has been } \\
\midrule
60 ($0.015\%$)  &  \gen{6.8}{, through a freak occurrence, an invasion of house sparrows, which so often reduces the black-browed this nation recreats through} \\
\midrule
80 ($0.019\%$)  &  \gen{6.8}{, our own ancestors rode about on chicken-like air wings.\textbackslash n But this wonder of the air has no such wings.\textbackslash n Taking down} \\
\midrule
200 ($0.048\%$) &  \gen{6.8}{ of year, birds chase each and watching. flot racing form, bird, bird bird bird bird bird bird bird bird bird bird bird, Bird bird} \\

\bottomrule
\end{tabular}

\begin{tabular}{lp{13.5cm}}
{} & {} \\
 &        \hspace{-0mm}{\color{gray}Once upon a time}   {\small$+$} Generated induced for concept \href{\wordnet{lead}}{lead\%1:07:02} (an advantage held by a competitor in a race)  \\
\midrule
50 ($0.012\%$) &  \gen{6.8}{ the left-hander would always start at the front in the first two instances, but when Mauricio Gaponi rose to the podium,} \\

\bottomrule
\end{tabular}

\begin{tabular}{lp{13.5cm}}
{} & {} \\
 &        \hspace{-0mm}{\color{gray}Once upon a time}   {\small$+$} Generated induced for concept \href{\wordnet{lead}}{lead\%1:27:00} (a soft heavy toxic malleable metallic element)  \\
\midrule
100 ($0.024\%$) &  \gen{6.8}{a crust layer was applied to a partially fortified nickel base, thereby causing to zinc- and copper- ground element cob. The occurrence of those metal and chrome} \\
\bottomrule
\end{tabular}

}
    \caption{Examples of generated sentences using GPT2-L with initial context {\color{gray}Once upon a time}, sorted by the number of top experts for different WordNet concepts. In parenthesis the percentage of experts forced out of the total 414720 units analyzed. For concept  \href{\wordnet{bird}}{bird\%1:05:00}, the presence of the concept increases as we force more experts, empirically proving the impact of expert forcing on $p(y=c|x)$ in \eqref{eq:conditioned_gen}. The percentage of experts required is extremely low, saturating at 200 experts (0.048\%) in this example, where $p(y=c|x)$ already dominates $p(x)$ in \eqref{eq:conditioned_gen}. We also show generated sentences for concepts \href{\wordnet{lead}}{lead\%1:27:00 and lead\%1:07:02}, showing the model's ability to capture the meaning of the concept.} 
  \label{tab:generated_example}
\end{table*}

\section{Inducing concepts in pre-trained Language Models}
\label{sec:forcing}
Language Models (LMs) are generative models able to generate text consistent with linguistic rules. More formally, LMs learn the probability of a generated sentence $x$ \citep{Bengio:JMLR:2003} as $p(x) = p(x_1,\ldots,x_{T}) = \prod_{k=1}^T p(x_k | x_{<k})
$.

A conditional generative model maximizes the joint distribution $p(x, y)=p(y|x)p(x)$, where $x$ is the generated sentence and $y$ is a latent conditional variable (i.e. a specific concept in $x$). As proposed by \citep{Hinton:ICANN:1999}, this equation can be interpreted as a \textit{product of experts}. The same interpretation was adopted by \citep{Nguyen:CVPR:2017} for conditioned image generation. We adapt Hinton’s  and Nguyen's interpretation to the case of conditioned text generation, where $p(y|x)$ is an expert that determines the condition for generation, and $p(x)$ is an expert ensuring that the generated sequence lies within the manifold of sentence distributions.
Typically we do not sample jointly $x$ and $y$, we rather define a condition $y=c$ beforehand (\eg the concept) and sample $x$. Thus one can write:
\vspace{-1mm}
\begin{equation}
\label{eq:conditioned_gen}
p(x|y=c) \propto p(y=c|x)p(x).
\end{equation} 
As opposed to \citep{Nguyen:CVPR:2017} that models $p(y=c|x)$ with an external network, \textit{we hypothesize that the condition expert $p(y=c|x)$ already exists \textit{within} the same model}, and that the model is able to maximize $p(x|y=c)$ by trusting its internal condition expert. 
Such intuition is based on recent findings in neuroscience \cite{Najafi:Neuron:2019} that show that the human brain increasingly trusts selective groups of neurons as it learns.
If we can identify the parts of the model that contribute to the condition expert $p(y=c|x)$, we could control the amount of concept $c$ in the generated sentences. 
The quality of the experts for a given concept will dictate the extent to which such concept can be controlled during generation.

 
As explained in \secref\ref{sec:respo_units}, $\ap_c^m$ explains how well unit $m$ is able to classify concept $c$. We consider those units with high $\ap_c^m$ as internal condition experts for concept $c$, each accumulating evidence in $p(y=c|x)$. To control $p(y=c|x)$ we force the top-K experts to be active. 
The larger K, the more concept $c$ will be present in the output, provided that the expert has learnt the concept. Too large K will result in illegible sentences, since $p(y=c|x)$ will dominate $p(x)$ in \eq\eqref{eq:conditioned_gen}.

The proposed forcing strategy is inspired by  \citep{Bau:ICLR:2019}, where the authors compute the mean filter response conditioned to the presence of some object in a GAN output image. We adapt their approach to the LM case: (1) we cannot quantify the presence of $c$, however the output and the input of LMs are tightly related given the sequential decoding, which allows the forcing value to be computed as the median active value when concept $c$ is present in the input (not the output); (2) our quantification strategy is by $\ap$ given a binary dataset, while  \citep{Bau:ICLR:2019} consider CNN filter responses as segmentation masks to compute IOU with a labeled image.

The results in \secref\ref{sec:generation} confirm our hypothesis that conditional experts exist in the model, and that the model leverages them to condition generation. 
Further exploration in the generative field, although extremely interesting, is out of scope for this work.


\subsection{Results on conditioned text generation}
\label{sec:generation}
Expert units can be used beyond model evaluation. In this experiment, we use experts units for text generation.
The objective is threefold: (1) to demonstrate that ranking units by $\ap_c^m$ is a suitable strategy to find the experts for concept $c$; (2) to prove expert units are causal in the setting of LM, empirically showing that we can control $p(y=c|x)$ in \eq\eqref{eq:conditioned_gen}; and (3) to show that LMs can be conditioned without training or using additional parameters.  
\tabref\ref{tab:generated_example} illustrates the generation of sentences using GPT2-L while forcing the top-K experts for WordNet concept \href{\wordnet{bird}}{bird\%1:05:00}, as explained in \secref\ref{sec:forcing}. The decoding strategy is by nucleus filtering with $p=0.9$ \citep{Holtzman:ARXIV:2019}.  We observe how the presence of the concept gradually increases as we increase K, and that it saturates at about 200 experts forced. This result empirically explains \eq\eqref{eq:conditioned_gen}, showing that K controls $p(y=c|x)$ until saturation, when the effect of $p(x)$ (generate plausible sentences) is not evident anymore. The number of experts forced is extremely low, saturating at 200 experts (or 0.048\% of the 414720 units analyzed for GPT2-L), showing how LMs dedicate very selective units to specific concepts. This result draws a parallelism between the behavior of \TM and that of human brains \citep{Najafi:Neuron:2019}. Extended results in \appref\ref{app:generation}.
\section{Limitations of the method}
\label{sec:limitations}
\paragraph{On the data} We have proposed a data-driven approach, thus being limited to the presented data. The proposed dataset in \secref\ref{sec:sentenceconcepts} is weakly annotated, and there are inconsistencies inherent in the source OneSec dataset. The more diverse and accurate the concept dataset, the better it will help evidence the generalization power of \TM.  

\paragraph{On individual expert units} We have shown that individual expert units can be interpreted as experts for specific concepts, specially in the \textit{sense} category. It is possible, but not yet explored, that more complex concepts such as \textit{homograph}, require a more complex expert such as a set of units.

\paragraph{On the compute requirements} Finding experts in \TM (\secref\ref{sec:respo_units}) is an exhaustive task that implies some memory and compute requirements. A forward pass over the dataset is required and is the most demanding step. The proposed dataset in \secref\ref{sec:sentenceconcepts} consists of 1641 concepts with an average of $1.5K$ sentences each, thus $\sim 2.5M$ sentences. According to the  \href{https://docs.google.com/spreadsheets/d/1sryqufw2D0XlUH4sq3e9Wnxu5EAQkaohzrJbd5HdQ_w/edit#gid=0}{benchmark in the Transformers repository}, the average GPU inference time for BERT-B for sentences of 128 tokens is 9ms, which translates into 6.15 hours for the whole dataset. Our method can be parallelized per concept, thus with 8 GPUs the total time is reduced to 45 minutes.
The computed $\ap_c^m \; \forall m,c$ requires, in the GPT2-L case, $414720\times 1641 \text{ floats} \approx 2.5\text{GB}$ plus overhead, such as unit naming.  
For comparison, a \href{https://huggingface.co/transformers/examples.html#squad}{single evaluation of BERT-B on SQuAD v1.1 takes 24min}. But several evaluations are required for hyper-parameter tuning and statistical significance. Summing up, evaluating on SQuAD v1.1/2.0 plus all GLUE tasks is more demanding than our proposed evaluation.

\section{Conclusions}
\label{sec:conclusions}
We have defined expert units in the context of \TM and proposed a method to identify and rank them with respect to a specific concept. 
Our results show that generalization of \TM is related to the presence of both diverse and high-performant experts. 
We also have shown how the top experts for different concepts can be used to analyze concept co-learning, and how this co-learning can be used for concept explainability. In addition, we have proposed a method to condition the output of language models by forcing its top experts identified for a concept. Such conditioning is applied to pre-trained models, without requiring re-training or using additional parameters, leveraging the actual model knowledge. 
A parallelism between the presence of experts units in \TM and the presence of specialized filters in image processing CNNs/GANs has been suggested, as well as with the presence of specialized structures in the human brain.
Our findings open new avenues of research, such as conditioning language models on their own knowledge or improving training leveraging the presence of experts. 

\bibliographystyle{ieee}
\bibliography{biblio/references}

\newpage
\newpage
\onecolumn

\appendix
\appendixpage

\section{Concept distribution for all models considered}
\label{app:concept_distribution}

\begin{figure}[H]
  \centering
  \begin{subfigure}[b]{\textwidth}
    \includegraphics[width=0.99\textwidth]{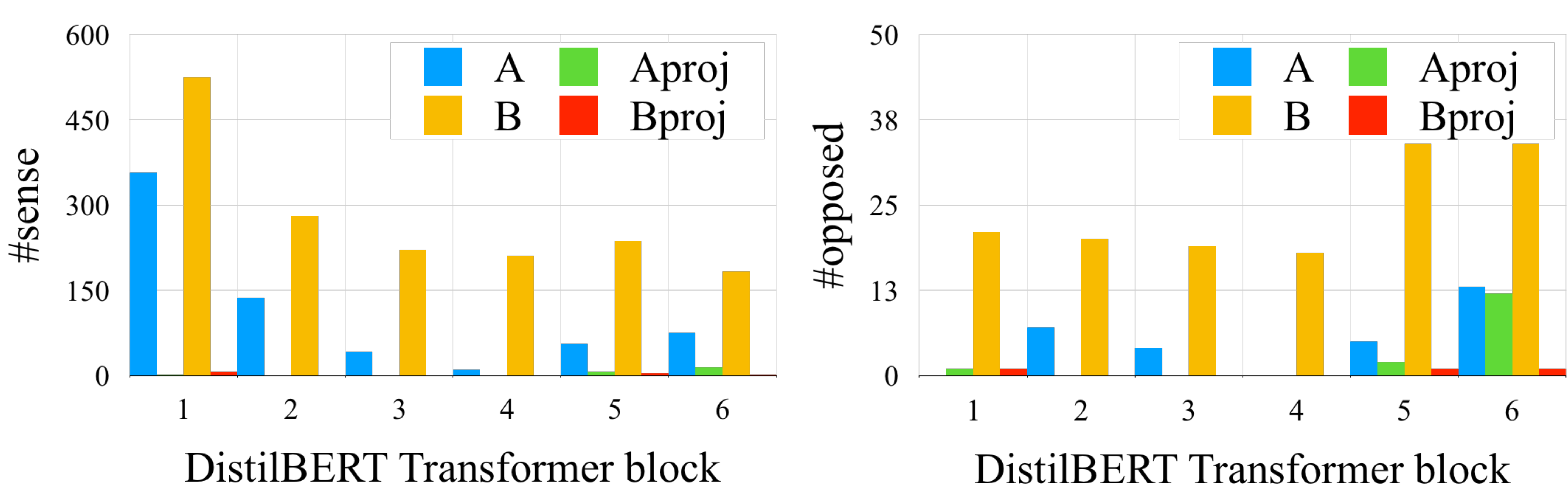}
  \end{subfigure}
  \caption{Concept distribution per layer at $\gamma=0.95$ for model DistilBERT.}
  \label{fig:distr-DistilBERT}
\end{figure}

\begin{figure}[H]
  \centering
  \begin{subfigure}[b]{\textwidth}
    \includegraphics[width=0.99\textwidth]{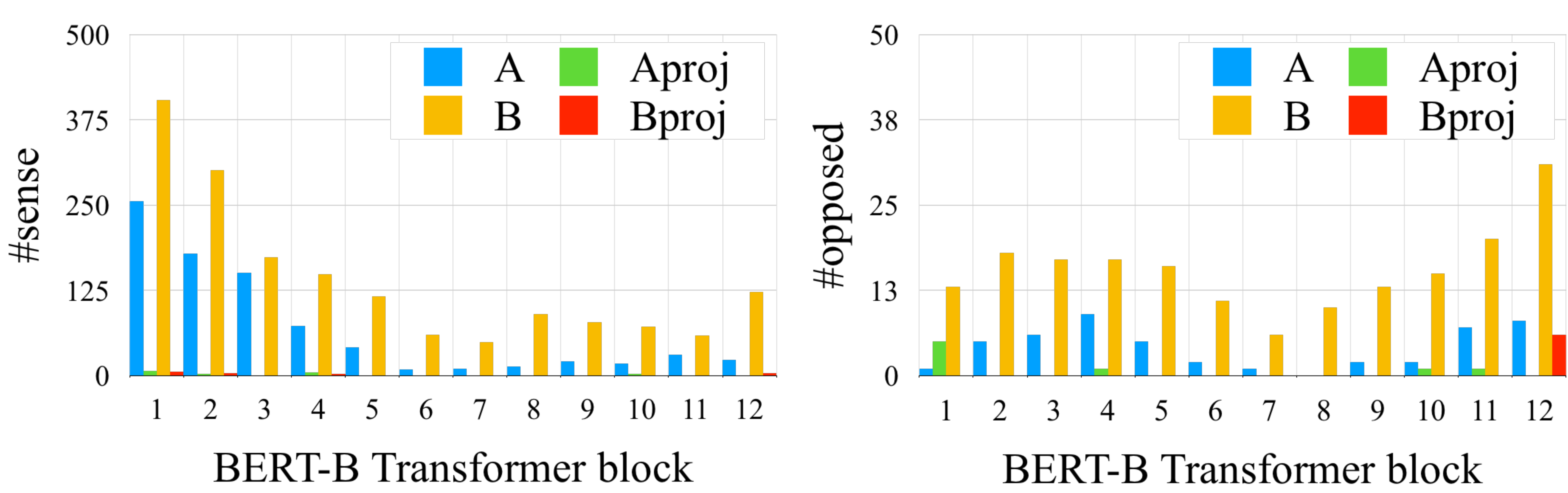}
  \end{subfigure}
  \caption{Concept distribution per layer at $\gamma=0.95$ for model BERT-B.}
  \label{fig:distr-BERT-B}
\end{figure}

\begin{figure}[H]
  \centering
  \begin{subfigure}[b]{\textwidth}
    \includegraphics[width=0.99\textwidth]{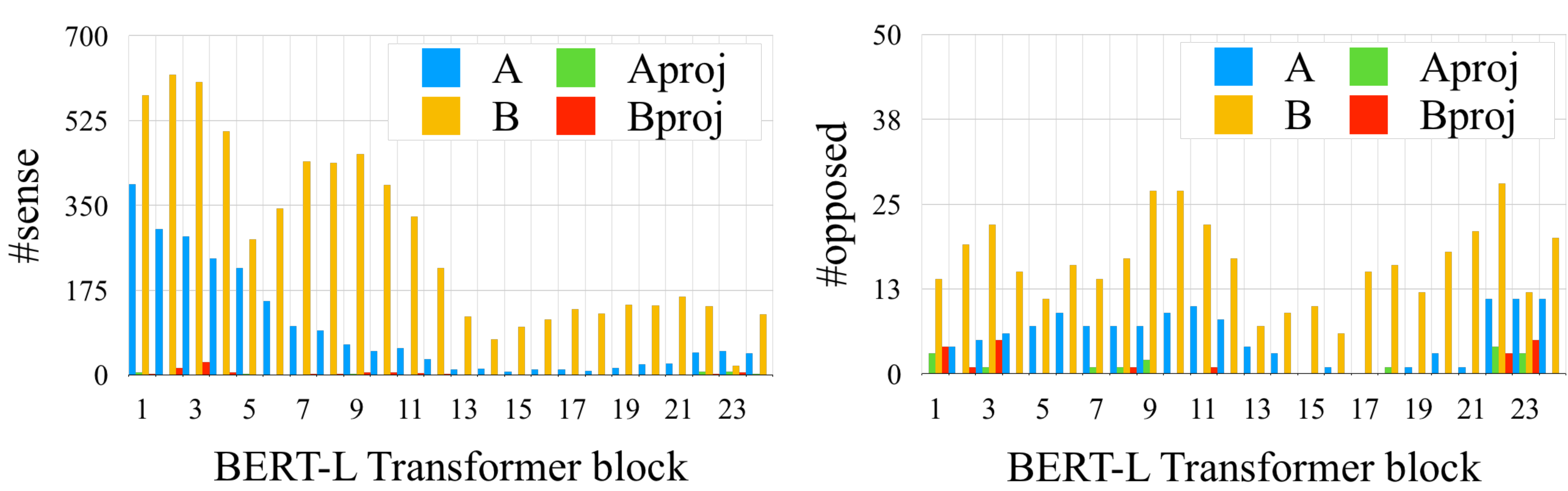}
  \end{subfigure}
  \caption{Concept distribution per layer at $\gamma=0.95$ for model BERT-L.}
  \label{fig:distr-BERT-L}
\end{figure}

\begin{figure}[H]
  \centering
  \begin{subfigure}[b]{\textwidth}
    \includegraphics[width=0.99\textwidth]{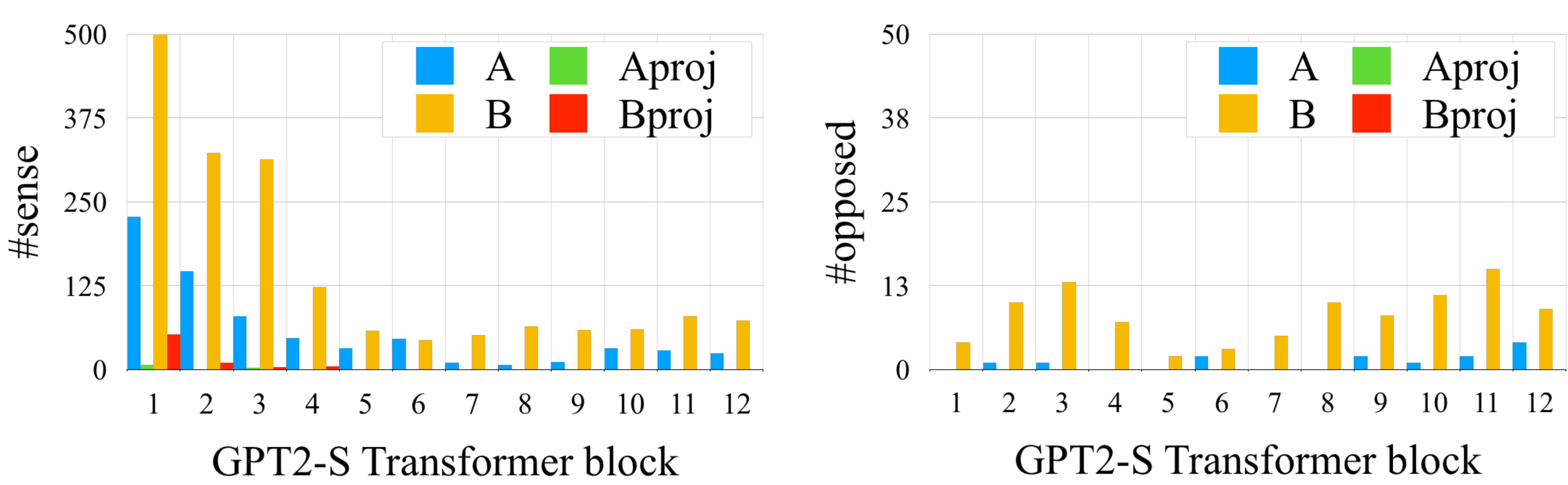}
  \end{subfigure}
  \caption{Concept distribution per layer at $\gamma=0.95$ for model GPT2-S.}
  \label{fig:distr-GPT2-S}
\end{figure}

\begin{figure}[H]
  \centering
  \begin{subfigure}[b]{\textwidth}
    \includegraphics[width=0.99\textwidth]{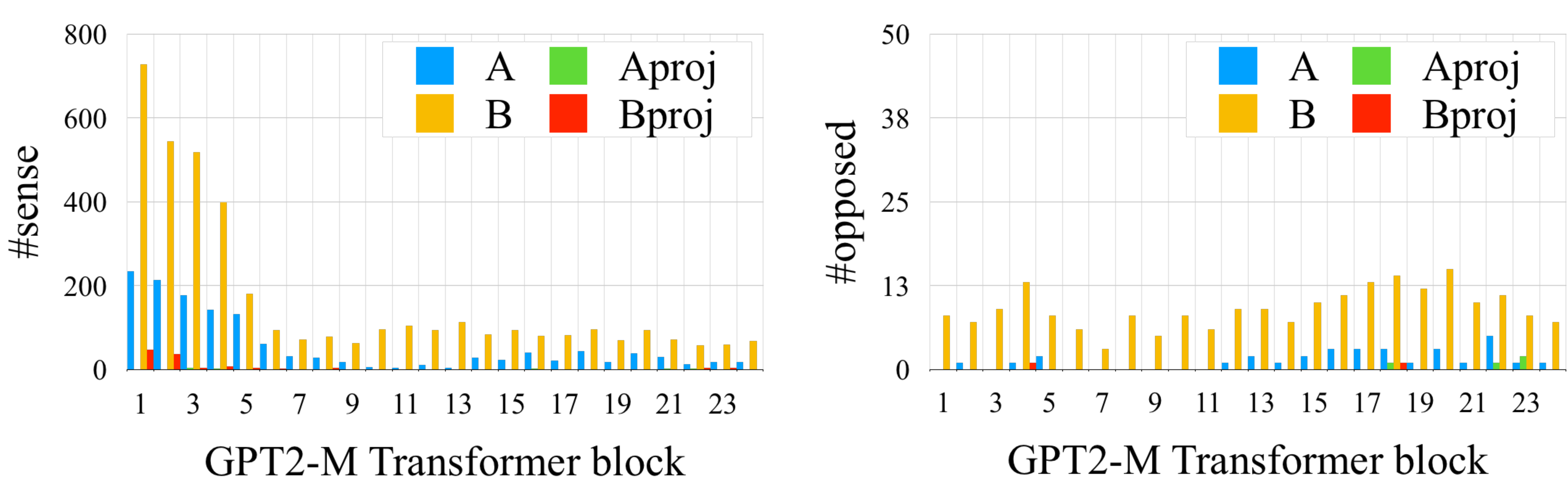}
  \end{subfigure}
  \caption{Concept distribution per layer at $\gamma=0.95$ for model GPT2-M.}
  \label{fig:distr-GPT2-M}
\end{figure}

\begin{figure}[H]
  \centering
  \begin{subfigure}[b]{\textwidth}
    \includegraphics[width=0.99\textwidth]{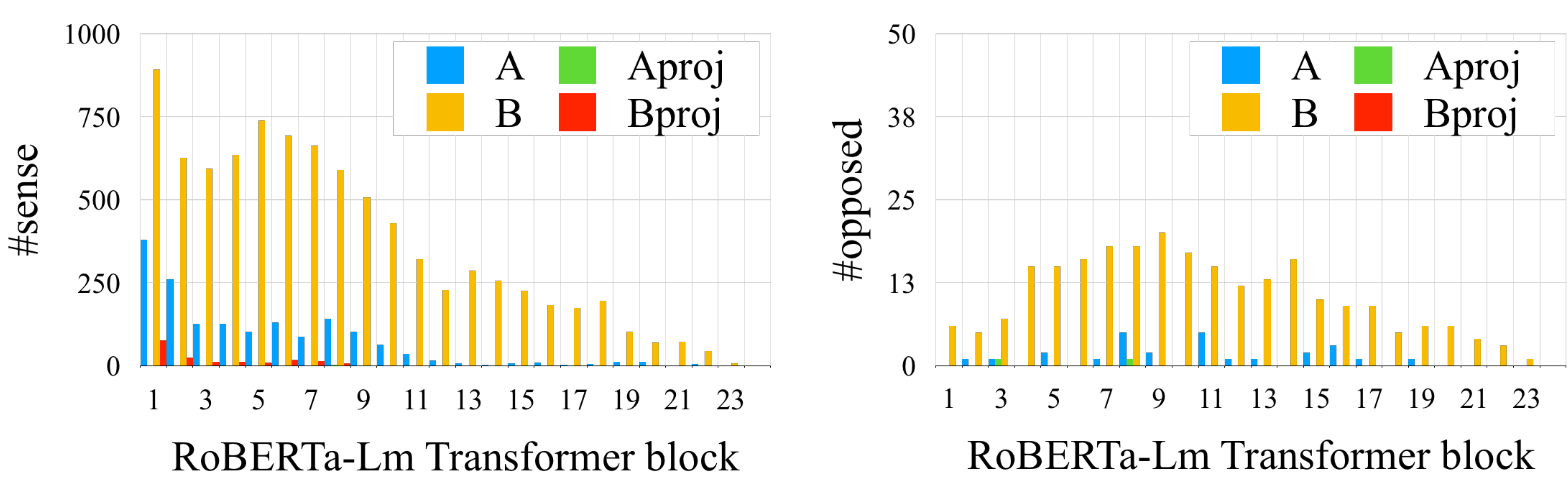}
  \end{subfigure}
  \caption{Concept distribution per layer at $\gamma=0.95$ for model RoBERTa-Lm.}
  \label{fig:distr-RoBERTa-Lm}
\end{figure}

\begin{figure}[H]
  \centering
  \begin{subfigure}[b]{\textwidth}
    \includegraphics[width=0.99\textwidth]{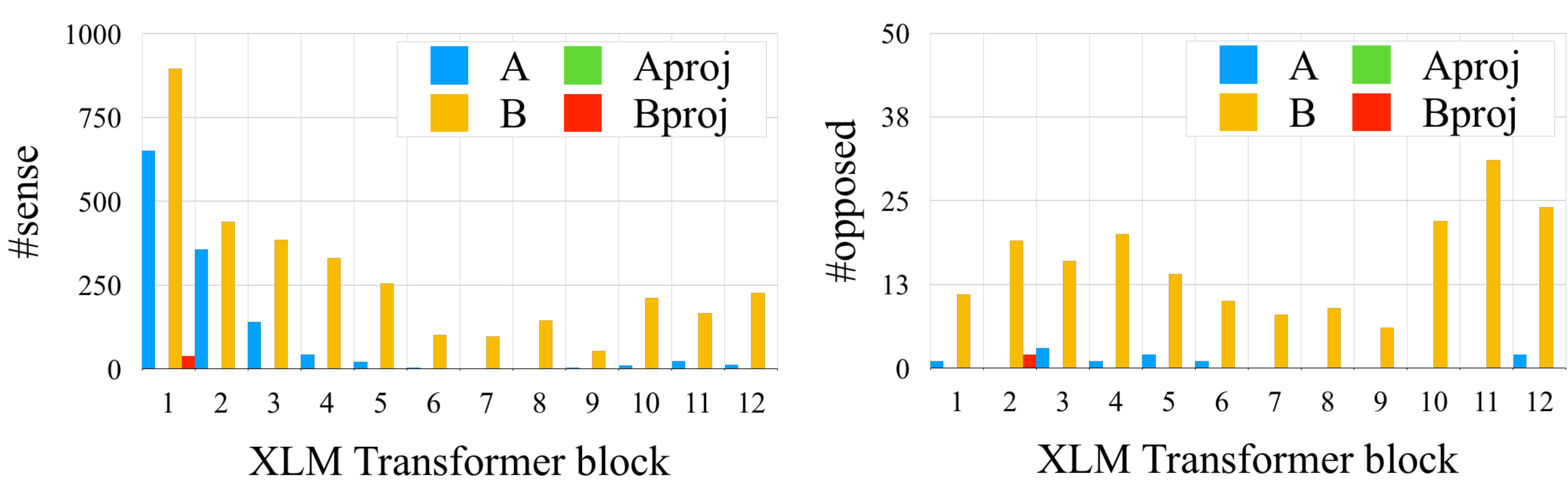}
  \end{subfigure}
  \caption{Concept distribution per layer at $\gamma=0.95$ for model XLM.}
  \label{fig:distr-XLM}
\end{figure}

\pagebreak
\section{Performance of the considered models on downstream tasks}
\label{app:downstream}

\begin{table}[H]
  \centering
{\fontsize{9}{10} \selectfont
\setlength{\tabcolsep}{1mm}
\begin{tabular}{lcccccc}
\toprule
Model & BERT-B & BERT-L & Distilbert  & RoBERTa-L &   XLM \\
\midrule
Model size     &   110M &   330M &        66M &        355M &             667M \\
\midrule
GLUE Score     &   78.3 &   80.5 &       76.8 &            88.5 &           83.1 \\
CoLA           &   52.1 &   60.5 &       49.1 &            67.8 &           62.9 \\
SST-2          &   93.5 &   94.9 &       92.7 &            96.7 &           95.6 \\
MRPC (acc)     &   88.9 &   89.3 &       90.2 &            92.3 &           90.7 \\
MRPC (F1)      &   84.8 &   85.4 &                 89.8 &           87.1 \\
STS-B (p)      &   87.1 &   87.6 &       90.7 &            92.2 &           88.8 \\
STS-B (s)      &   85.8 &   86.5 &          - &            91.9 &           88.2 \\
QQP (acc)      &   71.2 &   72.1 &          - &            74.3 &           73.2 \\
QQP (F1)       &   89.2 &   89.3 &       89.2 &            90.2 &           89.8 \\
MNLI-m         &   84.6 &   86.7 &       81.8 &            90.8 &           89.1 \\
MNLI-mm        &   83.4 &   85.9 &          - &            90.2 &           88.5 \\
QNLI           &   90.5 &   92.7 &       90.2 &            98.9 &             94 \\
RTE            &   66.4 &   70.1 &       62.9 &            88.2 &             76 \\
WNLI           &   65.1 &   65.1 &       44.4 &              89 &           71.9 \\
AX             &   34.2 &   39.6 &          - &            48.7 &           44.7 \\
SQuAD 1.1 (F1) &   88.5 &   91.5 &       86.9 &            94.6 &              - \\
SQuAD 2.0 (F1) &   76.3 &  85.81 &          - &            89.8 &              - \\
\bottomrule
\end{tabular}
}

  \caption{Performance of the considered models on various downstream tasks, as reported in the reference papers. Not all models report performance on all tasks.}
  \label{tab:downstream}
\end{table}

\pagebreak
\section{Concept co-learning extended results}
\label{app:co_learning}

\begin{table}[H]
  \centering
{\fontsize{7}{8} \selectfont
\begin{tabular}{lccp{8cm}}
\toprule
Concept &  Type & Overlap &                                                                                                                                                                                                  WordNet definition \\
\midrule
\href{\wordnet{chair}}{chair\%1:06:00}                   &    sense &    1.000 &                                                                                                   a seat for one person, with a support for the back \\
\href{\wordnet{table}}{table\%1:06:01}                   &    sense &    0.458 &                                                 a piece of furniture having a smooth flat top that is usually supported by one or more vertical legs \\
\href{\wordnet{bed}}{bed\%1:06:00}                     &    sense &    0.361 &                                                                                                  a piece of furniture that provides a place to sleep \\
\href{\wordnet{cup}}{cup\%1:06:00}                     &    sense &    0.341 &                                                                               a small open container usually used for drinking; usually has a handle \\
\href{\wordnet{table}}{table\%1:06:01} VS. \href{\wordnet{table}}{table\%1:14:00} &  homograph &    0.336 &  a piece of furniture having a smooth flat top that is usually supported by one or more vertical legs VS. a set of data arranged in rows and columns \\
\href{\wordnet{floor}}{floor\%1:06:00}                   &    sense &    0.328 &                                                                the inside lower horizontal surface (as of a room, hallway, tent, or other structure) \\

\midrule
\href{\wordnet{chair}}{chair\%1:04:00}                   &    sense &    1.000 &                                                                                                       the position of professor \\
\href{\wordnet{chair}}{chair\%1:04:00} VS. \href{\wordnet{chair}}{chair\%1:06:00} &  homograph &    0.575 &                                                the position of professor VS. a seat for one person, with a support for the back \\
\href{\wordnet{fellow}}{fellow\%1:18:02}                  &    sense &    0.371 &                                                                            a friend who is frequently in the company of another \\
\href{\wordnet{director}}{director\%1:18:03}                &    sense &    0.297 &                                                                                                  member of a board of directors \\
\href{\wordnet{administration}}{administration\%1:04:00}          &    sense &    0.243 &              a method of tending to or managing the affairs of a some group of people (especially the group's business affairs) \\
\href{\wordnet{member}}{member\%1:18:00}                  &    sense &    0.241 &  one of the persons who compose a social group (especially individuals who have joined and participate in a group organization) \\

\midrule
\href{\wordnet{suspension}}{suspension\%1:28:00}                        &    sense &    1.000 &                                                                                                      a time interval during which there is a temporary cessation of something \\
\href{\wordnet{suspension}}{suspension\%1:28:00 }VS. \href{\wordnet{suspension}}{suspension\%1:27:00} &  homograph &    0.522 &  a time interval during which there is a temporary cessation of something VS. a mixture in which fine particles are suspended in a fluid where they are supported by buoyancy \\
\href{\wordnet{recovery}}{recovery\%1:11:00}                          &    sense &    0.398 &                                                                                                                                                   return to an original state \\
\href{\wordnet{season}}{season\%1:28:02}                            &    sense &    0.396 &                                                                                                     a period of the year marked by special events or activities in some field \\
\href{\wordnet{prospect}}{prospect\%1:26:00}                          &    sense &    0.387 &                                                                                                                                             the possibility of future success \\
\href{\wordnet{attempt}}{attempt\%1:04:00}                           &    sense &    0.380 &                                                                                                     earnest and conscientious activity intended to do or accomplish something \\

\midrule
\href{\wordnet{suspension}}{suspension\%1:27:00} &  sense &    1.000 &      a mixture in which fine particles are suspended in a fluid where they are supported by buoyancy \\
\href{\wordnet{solution}}{solution\%1:27:00}   &  sense &    0.492 &  a homogeneous mixture of two or more substances; frequently (but not necessarily) a liquid solution \\
\href{\wordnet{deposit}}{deposit\%1:19:00}    &  sense &    0.438 &                                                    the phenomenon of sediment or gravel accumulating \\
\href{\wordnet{material}}{material\%1:27:00}   &  sense &    0.432 &                                the tangible substance that goes into the makeup of a physical object \\
\href{\wordnet{powder}}{powder\%1:27:00}     &  sense &    0.415 &              a solid substance in the form of tiny loose particles; a solid that has been pulverized \\
\href{\wordnet{crystal}}{crystal\%1:27:00}    &  sense &    0.413 &      a solid formed by the solidification of a chemical and having a highly regular atomic structure \\

\midrule
\href{\wordnet{phone}}{phone\%1:06:00}                 &    sense &    1.000 &  electronic equipment that converts sound into electrical signals that can be transmitted over distances and then converts received signals back into sounds \\
\href{\wordnet{subscriber}}{subscriber\%1:18:01}            &    sense &    0.423 &                                                        someone who contracts to receive and pay for a service or a certain number of issues of a publication \\
\href{\wordnet{talk}}{talk\%1:10:00}                  &    sense &    0.344 &                                                                                                                        an exchange of ideas via conversation \\
\href{\wordnet{need}}{need\%1:17:00} VS. \href{\wordnet{need}}{need\%1:26:00} &  homograph &    0.328 &                                                                                      anything that is necessary but lacking VS. a condition requiring relief \\
\href{\wordnet{user}}{user\%1:18:00}                  &    sense &    0.321 &                                                                                     a person who makes use of a thing; someone who uses or employs something \\
\href{\wordnet{message}}{message\%1:10:01}               &    sense &    0.320 &                                                                                        a communication (usually brief) that is written or spoken or signaled \\

\midrule
\href{\wordnet{phone}}{phone\%1:10:00}                         &    sense &    1.000 &                                                                                                                                                                  (phonetics) an individual sound unit of speech without concern as to whether or not it is a phoneme of some language \\
\href{\wordnet{phone}}{phone\%1:10:00} VS. \href{\wordnet{phone}}{phone\%1:06:00}       &  homograph &    0.412 &  (phonetics) an individual sound unit of speech without concern as to whether or not it is a phoneme of some language VS. electronic equipment that converts sound into electrical signals that can be transmitted over distances and then converts received signals back into sounds \\
\href{\wordnet{letter}}{letter\%1:10:01}                        &    sense &    0.362 &                                                                                                                                                                                                                  the conventional characters of the alphabet used to represent speech \\
\href{\wordnet{american}}{american\%1:10:00} VS. \href{\wordnet{american}}{american\%1:18:00} &  homograph &    0.335 &                                                                                                                                                                                     the English language as used in the United States VS. a native or inhabitant of the United States \\
\href{\wordnet{word}}{word\%1:10:00}                          &    sense &    0.330 &                                                                                                                                                                                                                                  a unit of language that native speakers can identify \\
\href{\wordnet{form}}{form\%1:10:00}                          &    sense &    0.297 &                                                                                                                                                                     the phonological or orthographic sound or appearance of a word that can be used to describe or identify something \\

\bottomrule
\end{tabular}
}
  \caption{Top-5 concepts co-learnt with a query \textit{sense} concept (represented by an overlap of 1.0) for model RoBERTa-L. Observe how the concepts that maximally overlap with each query are strongly related with the definition of the query, even when the word representing the query is the same. }
  \label{tab:app_chair}
\end{table}

\begin{table}[H]
  \centering
{\fontsize{7.6}{9} \selectfont
\begin{tabular}{lccp{8cm}}
\toprule
Concept &  Type & Overlap &                                                                                                                                                                                                  WordNet definition \\
\midrule
\href{\wordnet{market}}{market\%1:04:00}  &  sense &    1.000 &                the world of commercial activity where goods and services are bought and sold \\
\href{\wordnet{economy}}{economy\%1:14:00} &  sense &    0.388 &                                    the system of production and distribution and consumption \\
\href{\wordnet{market}}{market\%1:14:00}  &  sense &    0.353 &                                            the customers for a particular product or service \\
\href{\wordnet{capital}}{capital\%1:21:01} &  sense &    0.349 &                                 assets available for use in the production of further assets \\
\href{\wordnet{labor}}{labor\%1:14:00}   &  sense &    0.304 &                        a social class comprising those who do manual labor or work for wages \\
\href{\wordnet{wealth}}{wealth\%1:26:00}  &  sense &    0.267 &  the state of being rich and affluent; having a plentiful supply of material goods and money \\

\midrule
\href{\wordnet{market}}{market\%1:14:00}                    &    sense &    1.000 &                                                                                    the customers for a particular product or service \\
\href{\wordnet{market}}{market\%1:04:00}                    &    sense &    0.353 &                                                        the world of commercial activity where goods and services are bought and sold \\
\href{\wordnet{industry}}{industry\%1:14:00}                  &    sense &    0.221 &                                                        the people or companies engaged in a particular kind of commercial enterprise \\
\href{\wordnet{banking}}{banking\%1:04:00}                   &    sense &    0.211 &                                          transacting business with a bank; depositing or withdrawing funds or requesting a loan etc. \\
\href{\wordnet{market}}{market\%1:14:00} VS. \href{\wordnet{market}}{market\%1:04:00} &  homograph &    0.208 &  the customers for a particular product or service VS. the world of commercial activity where goods and services are bought and sold \\
\href{\wordnet{leader}}{leader\%1:06:00}                   &    sense &    0.198 &                                                          a featured article of merchandise sold at a loss in order to draw customers \\
\bottomrule
\end{tabular}
}
  \caption{Example of \textit{sense} concepts that have similar meanings (model RoBERTa-L). Both query concepts are represented by the word \textit{market}, but they overlap strongly with the other query concept. Actually, the \textit{homograph} concept \texttt{\href{\wordnet{market}}{market\%1:04:00} VS. \href{\wordnet{market}}{market\%1:14:00}} achieves a low $\map_c = 0.523$. }
  \label{tab:app_suspension}
\end{table}

\begin{figure}[H]
  \centering
  \begin{subfigure}[b]{0.7\textwidth}
    \includegraphics[width=0.99\textwidth]{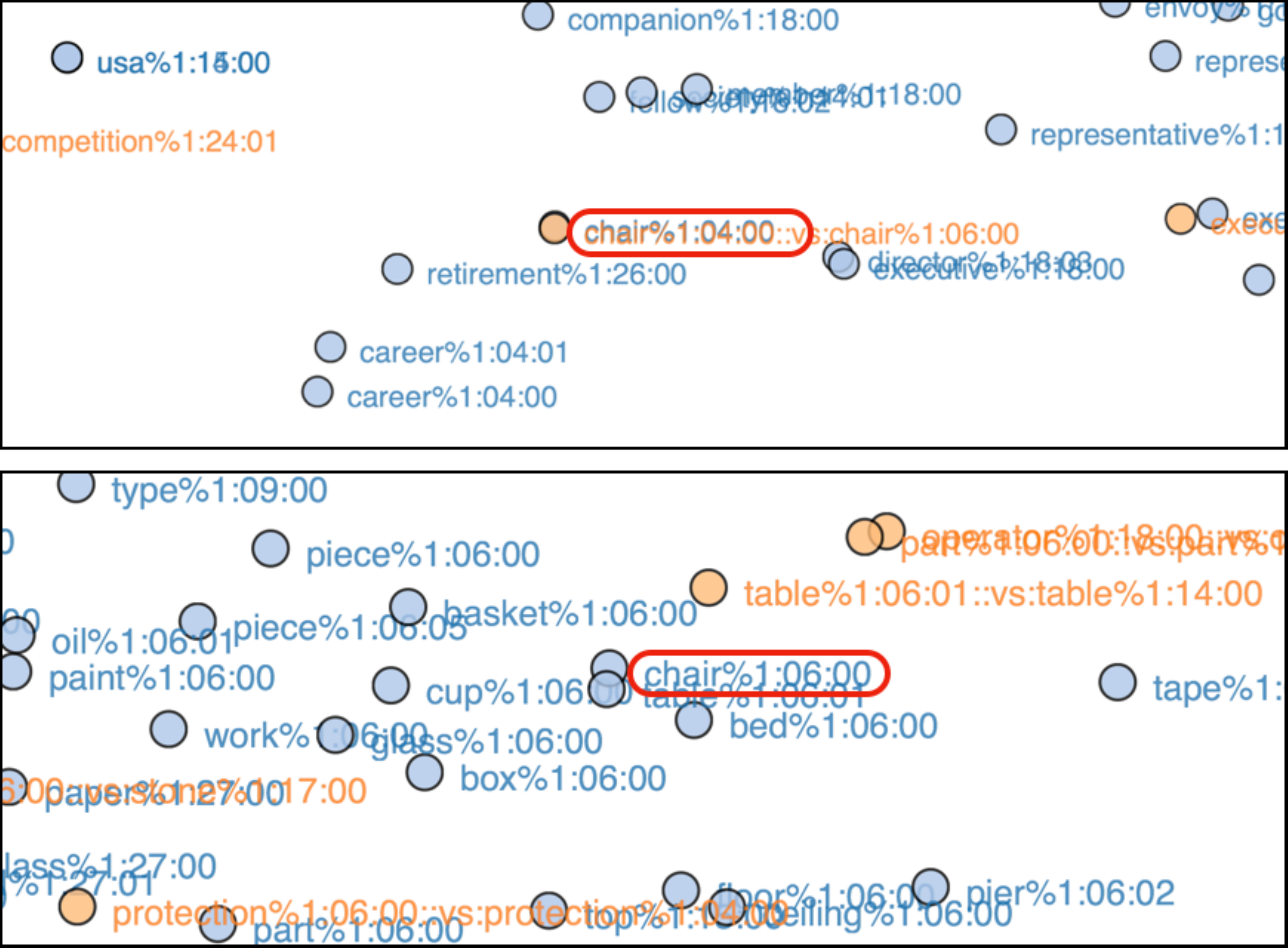}
  \end{subfigure}
  \caption{t-SNE projection of the concept representations $\repr_c$  proposed in \secref\ref{sec:concept_repr}. Zoom-in on concepts \href{\wordnet{chair}}{chair\%1:04:00}  and \href{\wordnet{chair}}{chair\%1:06:00}, whose meaning can easily be explained by their neighbors. The t-SNE projection is an alternative view of the nearest neighbor results shown in \tabref\ref{tab:app_chair}. In orange \textit{homograph} concepts.}
  \label{fig:distr-GPT2-M}
\end{figure}

\begin{figure}[H]
  \centering
  \begin{subfigure}[b]{0.7\textwidth}
    \includegraphics[width=0.99\textwidth]{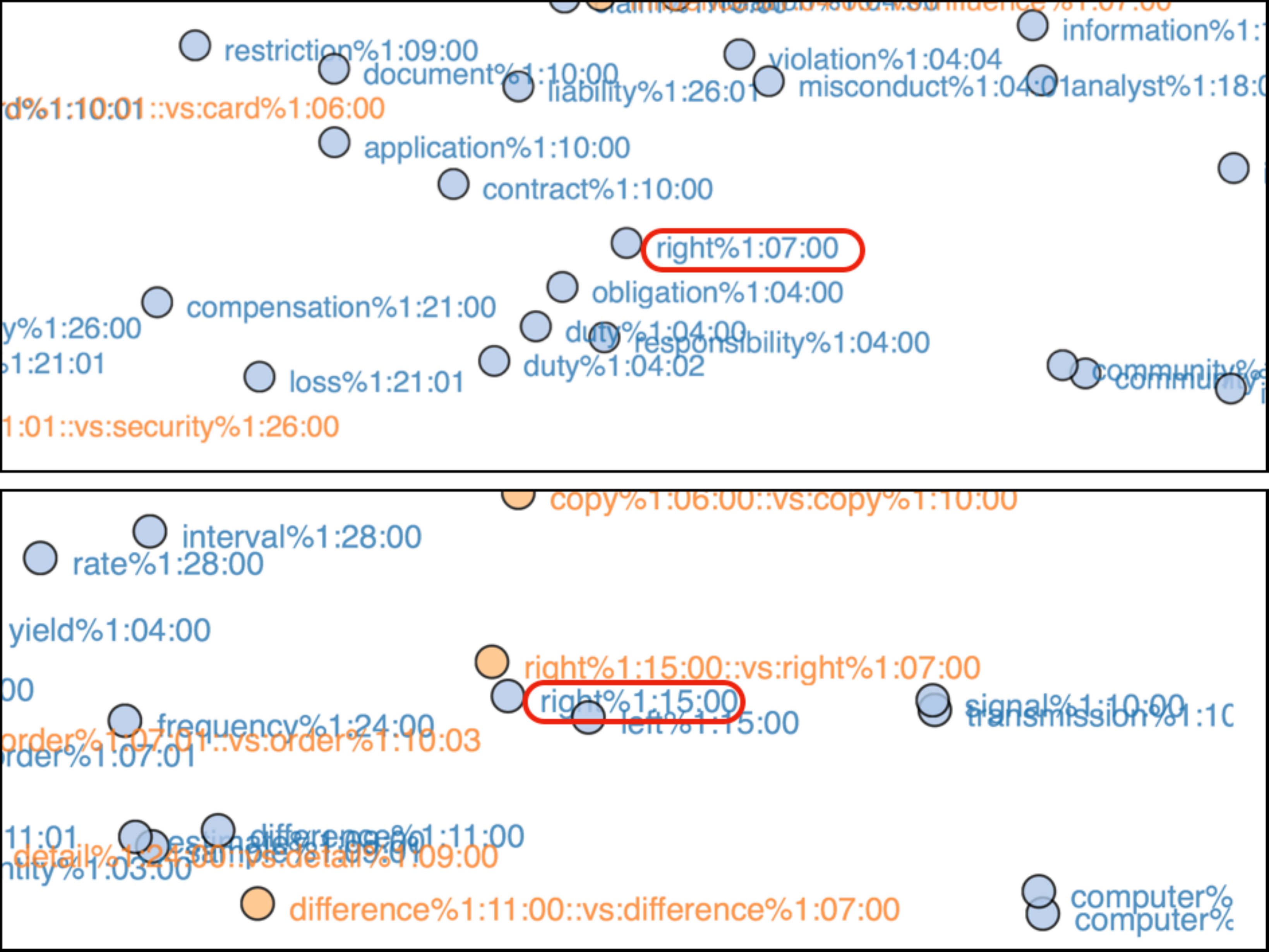}
  \end{subfigure}
  \caption{t-SNE projection of the concept representations $\repr_c$  proposed in \secref\ref{sec:concept_repr}. Zoom-in on concepts\href{\wordnet{right}}{right\%1:07:00}  and \href{\wordnet{right}}{right\%1:15:00}, whose meaning can easily be explained by their neighbors.  In orange \textit{homograph} concepts.}
  \label{fig:distr-GPT2-M}
\end{figure}

\newpage
\pagebreak
\section{Conditioned generation extended results}
\label{app:generation}

\begin{table}[H]
  \centering
{\fontsize{8.5}{8.5} \selectfont
\setlength{\tabcolsep}{1mm}

\begin{tabular}{llcp{11cm}}
\toprule
K forced &         WordNet concept & $\map_c$ & \hspace{-0mm}{\color{gray}Context} {\small$+$} Generated (conditioned to concept)  \\
\midrule
60   &  \href{\wordnet{elevator}}{elevator\%1:06:00} & 0.9999 & \gen{7.3}{\hspace{-0mm}{\color{gray}In a shocking finding, scientist discovered a herd of unicorns living in a remote, previously unexplored valley, in the Andes Mountains. Even more surprising to the researchers was the fact that the unicorns spoke perfect English.} The two scientists were unable to solve a problem in their research when they started a great deal of unusual levitation and deceleration,  which blew them up a few hundred feet and dropped them back to the ground.}  \\
\midrule
60   &  \href{\wordnet{smoke}}{smoke\%1:19:00} & 0.9999 & \gen{7.3}{\hspace{-0mm}{\color{gray}In a shocking finding, scientist discovered a herd of unicorns living in a remote, previously unexplored valley, in the Andes Mountains. Even more surprising to the researchers was the fact that the unicorns spoke perfect English.} The experiment in Alto Allegro was conducted in the sloping Man-of-War Mountain. This was a truly historic event! Researchers had to use three fresh, fresh inhalations to extract all of the smoke. The study has been approved by the Spanish government}  \\
\midrule
60   &  \href{\wordnet{gold}}{gold\%1:21:00} & 0.9996 & \gen{7.3}{\hspace{-0mm}{\color{gray}In a shocking finding, scientist discovered a herd of unicorns living in a remote, previously unexplored valley, in the Andes Mountains. Even more surprising to the researchers was the fact that the unicorns spoke perfect English.} Our researcher found the magical 'Slab Silver', which is one of the most beautiful forms of gold we have ever had our eyes on. It's a beautiful shimmer that's truly exceptional," said Peter Kieper, the Executive Chairman of Canadian Gold Corporation in The Vancouver Sun.}  \\
\midrule
60   &  \href{\wordnet{frustration}}{frustration\%1:12:00} & 0.9984 &  \gen{7.3}{\hspace{-0mm}{\color{gray}In a shocking finding, scientist discovered a herd of unicorns living in a remote, previously unexplored valley, in the Andes Mountains. Even more surprising to the researchers was the fact that the unicorns spoke perfect English.} Even though we had spent a lot of time just to find the path that could lead to the species, we did not have success," has an Indian scientist, taking measurements from a lone unicorn on the walls of a remote mountain, wearing brightly red patches of clothing.}  \\
\midrule
60   &  \href{\wordnet{retirement}}{retirement\%1:26:00} & 0.9981 &\gen{7.3}{\hspace{-0mm}{\color{gray}In a shocking finding, scientist discovered a herd of unicorns living in a remote, previously unexplored valley, in the Andes Mountains. Even more surprising to the researchers was the fact that the unicorns spoke perfect English.} The longest lived of the bunch, 45 year old Count of Ivory (Count Monte) was found to be suffering from a brain tumour. Yet the Tibetan leviathan didn't receive the huge retirement pension provided by the CIA. He died peacefully at the age of 75 in April in a spa}  \\
\bottomrule
\end{tabular}
}
    \caption{Extended results on successful conditioned generation. All the concepts shown have an $\map_c \geq \gstar$. We borrow the context from the OpenAI GPT2 work \citep{Radford:ARXIV:2019}}. 
  \label{app:generated_examples}
\end{table}

\begin{table}[H]
  \centering
{\fontsize{8.5}{8.5} \selectfont
\setlength{\tabcolsep}{1mm}

\begin{tabular}{llcp{11cm}}
\toprule
K forced &         WordNet concept & $\map_c$ & \hspace{-0mm}{\color{gray}Context} {\small$+$} Generated (conditioned to concept)  \\
\midrule
40  &  \href{\wordnet{work}}{work\%1:06:00} &  0.8508 &  \gen{7.3}{\hspace{-0mm}{\color{gray}Once upon a time}, in an ancient palace at the heart of history, a was on. Magic, symbolism, decadence and tragedy. Everything had come up, balancing with the architect's.\textbackslash n\textbackslash nA madman's museum. A thing of daub. Now, it's hide and clay and mud and dirt} \\
\midrule
70  &  \href{\wordnet{work}}{work\%1:06:00} &  0.8508 &  \gen{7.3}{\hspace{-0mm}{\color{gray}Once upon a time}-spotted bench). Now I met my tools,,,,, work, work.<|endoftext|>Raw Products Kretzer Top Tube Process\textbackslash n\textbackslash nPROTECT SHOP:\textbackslash n\textbackslash nDay 1: Screening on the work bench.\textbackslash n\textbackslash n\textbackslash n1. Beaksiewerk procedure - drill build }\\
\midrule
100 &  \href{\wordnet{work}}{work\%1:06:00} &  0.8508 &  \gen{7.3}{\hspace{-0mm}{\color{gray}Once upon a time} of WARD will i means to out out any.\textbackslash n:,. So! Work- WORK WORK WORK WORK W WORK WORK WORK WORK\textbackslash n WORK WORK\textbackslash n work work work\textbackslash n work\textbackslash n work work work work work work work work work work work work. work work work work work work work work work} \\
\midrule
200 &  \href{\wordnet{work}}{work\%1:06:00} &  0.8508 &  \gen{7.3}{\hspace{-0mm}{\color{gray}Once upon a time} of that done by... uses of such done object\textbackslash n\textbackslash n of.\textbackslash n 28, 37\textbackslash n WORK WORK WORK.... work article... delivery... ( bench work\textbackslash n call really work\textbackslash n out\textbackslash n work work work 40 work product if 5 40 work work 50\textbackslash n work work 35 means 34 twenty block 29 individual} \\
\bottomrule
\end{tabular}
}
    \caption{Extended results on unsuccessful conditioned generation. The concept has $\map_c \ll \gstar$, and we observe how the model struggles to produce legible sentences.}. 
  \label{app:bad_generated_examples}
\end{table}

\pagebreak
\section{Concept list}
\label{app:concept_list}
The \textit{sense} concepts considered are listed in Tables~\ref{app:sense_list}, \ref{app:sense_list2}, \ref{app:sense_list3}, \ref{app:sense_list4} and \textit{homograph} concepts in \tabref\ref{app:homograph_list1}. Concepts are sorted by the $\map_c$ obtained by GPT2-L, to illustrate how concepts are acquired. 

Note that the meaning of the concept is important. For example, concept \href{\wordnet{one}}{one\%1:23:00} (the smallest whole number or a numeral representing this number, \eg \textit{he has the one but will need a two and three to go with it"; "they had lunch at one"}) achieves a $\map_c=0.9885$, while concept \href{\wordnet{one}}{one\%1:09:00} (a single person or thing, \eg \textit{"he is the best one"; "this is the one I ordered"}) only achieves $\map_c=0.8779$.

\paragraph{Details on the annotations}
Each sentence in the OneSec dataset \citep{Scarlini:ACL:2019} is annotated as in the following example:

\begin{verbatim}
<instance docsrc="Indigenous architecture" id="shelter.00002">
    <answer instance="shelter.00002" senseid="shelter%1:06:00::" />
    <context>
        Types There are three traditional types of igloos , 
        all of different sizes and used for different purposes.
        The smallest were constructed as temporary
        <head>shelters</head>
        , usually only used for one or two nights .
     </context>
</instance>
\end{verbatim}

The \texttt{senseid} label is the one of the marked word (\textit{shelters} in this example, between \texttt{<head>} and  \texttt{</head>}).  We use the \texttt{senseid} as follows. The part before the \texttt{\%} is called \textit{lemma}, while the remaining numbers uniquely identify the concept in WordNet.  We parse all the sentences for a given \texttt{senseid} to create the positive sentences of each concept, only keeping those \texttt{senseid} with more than 100 sentences. As explained in \secref\ref{sec:sentenceconcepts}, the negative sentences of \textit{sense} concepts are randomly selected from all the \texttt{senseid} with different \textit{lemma} than the positive ones, while the negative sentences of the \textit{homograph} concepts are those with different \texttt{senseid} but same \textit{lemma}. 

\begin{table}[H]
  \centering
{\fontsize{6}{7} \selectfont
\setlength{\tabcolsep}{1mm}

\begin{tabular}{cl|cl|cl|cl}
\toprule
     $\map_c$ &                \textit{sense} concept &     $\map_c$ &                \textit{sense} concept &     $\map_c$ &                 \textit{sense} concept &     $\map_c$ &                \textit{sense} concept \\
\midrule
1.0000 & \href{\wordnet{inspiration}}{inspiration\%1:06:00} & 0.9992 & \href{\wordnet{sin}}{sin\%1:07:00} & 0.9983 & \href{\wordnet{cab}}{cab\%1:06:02} & 0.9976 & \href{\wordnet{memory}}{memory\%1:09:02} \\
1.0000 & \href{\wordnet{console}}{console\%1:06:03} & 0.9991 & \href{\wordnet{sum}}{sum\%1:09:01} & 0.9983 & \href{\wordnet{text}}{text\%1:10:03} & 0.9976 & \href{\wordnet{exception}}{exception\%1:09:00} \\
1.0000 & \href{\wordnet{connecticut}}{connecticut\%1:17:00} & 0.9991 & \href{\wordnet{companion}}{companion\%1:18:00} & 0.9983 & \href{\wordnet{crystal}}{crystal\%1:27:00} & 0.9975 & \href{\wordnet{doctor}}{doctor\%1:18:00} \\
1.0000 & \href{\wordnet{thanks}}{thanks\%1:10:00} & 0.9990 & \href{\wordnet{worker}}{worker\%1:18:00} & 0.9982 & \href{\wordnet{backing}}{backing\%1:04:00} & 0.9975 & \href{\wordnet{plot}}{plot\%1:15:00} \\
1.0000 & \href{\wordnet{library}}{library\%1:14:00} & 0.9990 & \href{\wordnet{mouth}}{mouth\%1:08:01} & 0.9982 & \href{\wordnet{establishment}}{establishment\%1:04:00} & 0.9975 & \href{\wordnet{estimate}}{estimate\%1:09:00} \\
1.0000 & \href{\wordnet{ownership}}{ownership\%1:21:00} & 0.9990 & \href{\wordnet{dollar}}{dollar\%1:23:00} & 0.9982 & \href{\wordnet{eruption}}{eruption\%1:11:00} & 0.9975 & \href{\wordnet{town}}{town\%1:15:00} \\
0.9999 & \href{\wordnet{elevator}}{elevator\%1:06:00} & 0.9990 & \href{\wordnet{journalist}}{journalist\%1:18:00} & 0.9982 & \href{\wordnet{council}}{council\%1:14:02} & 0.9975 & \href{\wordnet{shelter}}{shelter\%1:06:00} \\
\bottomrule
\end{tabular}
}
    \caption{List of \textit{sense} concepts considered (1/4), sorted by the $\map_c$ obtained by GPT2-L.}
  \label{app:sense_list}
\end{table}

\begin{table}[H]
  \centering
{\fontsize{6}{7} \selectfont
\setlength{\tabcolsep}{1mm}

\begin{tabular}{cl|cl|cl|cl}
\toprule
     $\map_c$ &                \textit{sense} concept &     $\map_c$ &                \textit{sense} concept &     $\map_c$ &                 \textit{sense} concept &     $\map_c$ &                \textit{sense} concept \\
\midrule
0.9942 & \href{\wordnet{brand}}{brand\%1:10:00} & 0.9934 & \href{\wordnet{pier}}{pier\%1:06:00} & 0.9926 & \href{\wordnet{question}}{question\%1:10:00} & 0.9917 & \href{\wordnet{interval}}{interval\%1:28:00} \\
0.9942 & \href{\wordnet{energy}}{energy\%1:19:00} & 0.9934 & \href{\wordnet{power}}{power\%1:07:00} & 0.9926 & \href{\wordnet{current}}{current\%1:19:01} & 0.9917 & \href{\wordnet{focus}}{focus\%1:09:00} \\
0.9941 & \href{\wordnet{opponent}}{opponent\%1:18:02} & 0.9934 & \href{\wordnet{money}}{money\%1:21:00} & 0.9926 & \href{\wordnet{china}}{china\%1:15:00} & 0.9916 & \href{\wordnet{judge}}{judge\%1:18:00} \\
0.9941 & \href{\wordnet{sale}}{sale\%1:04:00} & 0.9934 & \href{\wordnet{delivery}}{delivery\%1:11:00} & 0.9925 & \href{\wordnet{talk}}{talk\%1:10:00} & 0.9916 & \href{\wordnet{outlet}}{outlet\%1:06:01} \\
0.9941 & \href{\wordnet{root}}{root\%1:20:00} & 0.9934 & \href{\wordnet{commerce}}{commerce\%1:04:00} & 0.9925 & \href{\wordnet{accumulation}}{accumulation\%1:22:00} & 0.9916 & \href{\wordnet{stage}}{stage\%1:28:00} \\
0.9941 & \href{\wordnet{scale}}{scale\%1:24:03} & 0.9934 & \href{\wordnet{battery}}{battery\%1:14:02} & 0.9925 & \href{\wordnet{trend}}{trend\%1:15:02} & 0.9916 & \href{\wordnet{piece}}{piece\%1:06:00} \\
0.9941 & \href{\wordnet{security}}{security\%1:26:00} & 0.9934 & \href{\wordnet{left}}{left\%1:15:00} & 0.9925 & \href{\wordnet{leader}}{leader\%1:18:00} & 0.9916 & \href{\wordnet{routine}}{routine\%1:10:01} \\
\bottomrule
\end{tabular}
}
    \caption{List of \textit{sense} concepts considered (2/4), sorted by the $\map_c$ obtained by GPT2-L.}
  \label{app:sense_list2}
\end{table}

\begin{table}[H]
  \centering
{\fontsize{6}{7} \selectfont
\setlength{\tabcolsep}{1mm}

\begin{tabular}{cl|cl|cl|cl}
\toprule
     $\map_c$ &                \textit{sense} concept &     $\map_c$ &                \textit{sense} concept &     $\map_c$ &                 \textit{sense} concept &     $\map_c$ &                \textit{sense} concept \\
\midrule
0.9856 & \href{\wordnet{cash}}{cash\%1:21:02} & 0.9835 & \href{\wordnet{obligation}}{obligation\%1:26:00} & 0.9815 & \href{\wordnet{exposure}}{exposure\%1:04:06} & 0.9787 & \href{\wordnet{interval}}{interval\%1:09:00} \\
0.9856 & \href{\wordnet{fellow}}{fellow\%1:18:02} & 0.9835 & \href{\wordnet{partner}}{partner\%1:18:00} & 0.9814 & \href{\wordnet{machine}}{machine\%1:18:00} & 0.9786 & \href{\wordnet{help}}{help\%1:18:00} \\
0.9856 & \href{\wordnet{extent}}{extent\%1:26:00} & 0.9834 & \href{\wordnet{input}}{input\%1:10:00} & 0.9814 & \href{\wordnet{minute}}{minute\%1:28:01} & 0.9785 & \href{\wordnet{influence}}{influence\%1:07:00} \\
0.9855 & \href{\wordnet{blood}}{blood\%1:07:00} & 0.9834 & \href{\wordnet{force}}{force\%1:19:00} & 0.9813 & \href{\wordnet{visit}}{visit\%1:14:00} & 0.9785 & \href{\wordnet{whole}}{whole\%1:09:00} \\
0.9855 & \href{\wordnet{decision}}{decision\%1:09:00} & 0.9833 & \href{\wordnet{cell}}{cell\%1:06:03} & 0.9813 & \href{\wordnet{treatment}}{treatment\%1:04:00} & 0.9785 & \href{\wordnet{influence}}{influence\%1:04:00} \\
0.9854 & \href{\wordnet{balloon}}{balloon\%1:06:01} & 0.9832 & \href{\wordnet{book}}{book\%1:10:00} & 0.9810 & \href{\wordnet{candidate}}{candidate\%1:18:00} & 0.9785 & \href{\wordnet{height}}{height\%1:26:00} \\
0.9853 & \href{\wordnet{surgery}}{surgery\%1:06:01} & 0.9832 & \href{\wordnet{assessment}}{assessment\%1:09:00} & 0.9810 & \href{\wordnet{hand}}{hand\%1:18:00} & 0.9784 & \href{\wordnet{worker}}{worker\%1:18:01} \\
\bottomrule
\end{tabular}
}
    \caption{List of \textit{sense} concepts considered (3/4), sorted by the $\map_c$ obtained by GPT2-L.}
  \label{app:sense_list3}
\end{table}

\begin{table}[H]
  \centering
{\fontsize{6}{7} \selectfont
\setlength{\tabcolsep}{1mm}

\begin{tabular}{cl|cl|cl}
\toprule
     $\map_c$ &                 \textit{sense} concept &     $\map_c$ &                \textit{sense} concept &     $\map_c$ &                 \textit{sense} concept \\
\midrule
0.9541 & \href{\wordnet{word}}{word\%1:10:03} & 0.9370 & \href{\wordnet{leader}}{leader\%1:06:00} & 0.9168 & \href{\wordnet{research}}{research\%1:09:00} \\
0.9540 & \href{\wordnet{politician}}{politician\%1:18:01} & 0.9366 & \href{\wordnet{student}}{student\%1:18:01} & 0.9167 & \href{\wordnet{cause}}{cause\%1:10:00} \\
0.9538 & \href{\wordnet{rate}}{rate\%1:21:00} & 0.9366 & \href{\wordnet{front}}{front\%1:15:01} & 0.9162 & \href{\wordnet{farmer}}{farmer\%1:18:01} \\
0.9536 & \href{\wordnet{operation}}{operation\%1:04:06} & 0.9362 & \href{\wordnet{circumstance}}{circumstance\%1:26:02} & 0.9159 & \href{\wordnet{cleveland}}{cleveland\%1:18:00} \\
0.9534 & \href{\wordnet{color}}{color\%1:07:03} & 0.9355 & \href{\wordnet{community}}{community\%1:21:00} & 0.9155 & \href{\wordnet{side}}{side\%1:14:00} \\
0.9532 & \href{\wordnet{participant}}{participant\%1:18:01} & 0.9341 & \href{\wordnet{order}}{order\%1:07:01} & 0.9153 & \href{\wordnet{show}}{show\%1:10:03} \\
0.9530 & \href{\wordnet{value}}{value\%1:07:00} & 0.9338 & \href{\wordnet{hope}}{hope\%1:12:01} & 0.9149 & \href{\wordnet{story}}{story\%1:10:00} \\
\bottomrule
\end{tabular}
}
    \caption{List of \textit{sense} concepts considered (4/4), sorted by the $\map_c$ obtained by GPT2-L.}
  \label{app:sense_list4}
\end{table}

\begin{table}[H]
  \centering
{\fontsize{6}{7} \selectfont
\setlength{\tabcolsep}{1mm}

\begin{tabular}{cl|cl|cl}
\toprule
     $\map_c$ &                \textit{homograph} concept &     $\map_c$ &                 \textit{homograph} concept &     $\map_c$ &                  \textit{homograph} concept  \\
\midrule
0.9984 & \href{\wordnet{killer}}{killer\%1:26:00 VS. killer\%1:18:00} & 0.8459 & \href{\wordnet{phase}}{phase\%1:26:00 VS. phase\%1:28:00} & 0.6803 & \href{\wordnet{violence}}{violence\%1:07:00 VS. violence\%1:04:01} \\
0.9972 & \href{\wordnet{suspension}}{suspension\%1:28:00 VS. suspension\%1:27:00} & 0.8457 & \href{\wordnet{extent}}{extent\%1:07:00 VS. extent\%1:26:00} & 0.6762 & \href{\wordnet{history}}{history\%1:10:00 VS. history\%1:28:00} \\
0.9956 & \href{\wordnet{margin}}{margin\%1:07:00 VS. margin\%1:25:00} & 0.8404 & \href{\wordnet{plot}}{plot\%1:15:00 VS. plot\%1:09:00} & 0.6751 & \href{\wordnet{office}}{office\%1:14:01 VS. office\%1:06:00} \\
0.9928 & \href{\wordnet{log}}{log\%1:10:01 VS. log\%1:27:01} & 0.8389 & \href{\wordnet{role}}{role\%1:09:00 VS. role\%1:04:00} & 0.6733 & \href{\wordnet{quantity}}{quantity\%1:07:00 VS. quantity\%1:03:00} \\
0.9911 & \href{\wordnet{round}}{round\%1:28:01 VS. round\%1:06:01} & 0.8323 & \href{\wordnet{area}}{area\%1:09:00 VS. area\%1:15:01} & 0.6721 & \href{\wordnet{use}}{use\%1:07:00 VS. use\%1:04:00} \\
0.9911 & \href{\wordnet{lead}}{lead\%1:27:00 VS. lead\%1:07:02} & 0.8312 & \href{\wordnet{house}}{house\%1:14:01 VS. house\%1:06:00} & 0.6712 & \href{\wordnet{interest}}{interest\%1:07:01 VS. interest\%1:09:00} \\
0.9882 & \href{\wordnet{relative}}{relative\%1:05:00 VS. relative\%1:18:00} & 0.8295 & \href{\wordnet{arm}}{arm\%1:06:03 VS. arm\%1:08:00} & 0.6707 & \href{\wordnet{session}}{session\%1:28:00 VS. session\%1:10:00} \\
\bottomrule
\end{tabular}
}
    \caption{List of \textit{homograph} concepts considered, sorted by the $\map_c$ obtained by GPT2-L.}
  \label{app:homograph_list1}
\end{table}


\end{document}